\documentclass[journal]{IEEEtran}
\usepackage{epsfig}
\usepackage{subfigure}
\usepackage{amsmath}
\usepackage{amssymb}
\usepackage{makecell}
\usepackage{booktabs}
\usepackage[switch]{lineno}
\usepackage{amsmath,algorithm,algpseudocode,amsfonts,algorithmicx}

\usepackage{color}

\ifCLASSINFOpdf
\else
\fi
\begin{document}
\title{RSDet++: Point-based Modulated Loss for More Accurate Rotated Object Detection}

%\author{Michael~Shell,~\IEEEmembership{Member,~IEEE,}
%        John~Doe,~\IEEEmembership{Fellow,~OSA,}
%        and~Jane~Doe,~\IEEEmembership{Life~Fellow,~IEEE}% <-this % stops a space
%\thanks{M. Shell was with the Department
%of Electrical and Computer Engineering, Georgia Institute of Technology, Atlanta,

\author{    Wen Qian,
    Xue Yang, 
    Silong Peng,  
    Junchi Yan ~\IEEEmembership{Member,~IEEE,}
    Xiujuan Zhang
    \thanks{Wen Qian, Silong peng (Corresponding author) are with Institute of Automation, Chinese Academy of Sciences,
    and also with the University of Chinese Academy of Sciences.
    Xue Yang, Junchi Yan are with Department of Computer Science and Engineering, Shanghai Jiao Tong University.
    Xiujuan Zhang is with the Inner Mongolia Key Laboratory of Molecular Biology on Featured Plants.}
    \thanks{E-mail: \{qianwen2018, silong.peng\}@ia.ac.cn, \{yangxue-2019-sjtu, yanjunchi\}@sjtu.edu.cn, mingyuesong@163.com}
    }
%\thanks{M. Shell was with the Department
%of Electrical and Computer Engineering, Georgia Institute of Technology, Atlanta,

\markboth{Journal of \LaTeX\ Class Files,~Vol.~14, No.~8, August~2015}%
{Shell \MakeLowercase{\textit{et al.}}: Bare Demo of IEEEtran.cls for IEEE Journals}



\maketitle

\begin{abstract}
%Popular rotated detection methods usually use five parameters (coordinates of the central point, width, height, and rotation angle) or eight parameters (coordinates of four vertices) to describe the rotated bounding box and $\ell_1$ loss as the loss function. 
%In this paper, we argue that the aforementioned integration can cause training instability and performance degeneration. 
%The main reason is the discontinuity of loss which is caused by the contradiction between the definition of the rotated bounding box and the loss function. 
%We refer to the above issues as rotation sensitivity error (RSE) and propose a modulated rotation loss to dismiss the discontinuity of loss. 
%The modulated rotation loss can achieve consistent improvement on the five parameter methods and the eight parameter methods. 
%Experimental results using one stage and two stages detectors demonstrate the effectiveness of our loss. 
%The integrated network achieves competitive performances on several benchmarks including DOTA and UCAS AOD. The code is available at https://github.com/yangxue0827/RotationDetection.

We classify the discontinuity of loss in both five-param and eight-param rotated object detection methods as rotation sensitivity error (RSE) 
which will result in performance degeneration.
We introduce a novel modulated rotation loss to alleviate the problem and 
propose a rotation sensitivity detection network (RSDet) which is consists of an eight-param single-stage rotated object detector and the modulated rotation loss.
Our proposed RSDet has several advantages:
1) it reformulates the rotated object detection problem as predicting the corners of objects while most previous methods employ a five-para-based regression method with different measurement units.
2) modulated rotation loss achieves consistent improvement on both five-param and eight-param rotated object detection methods by solving the discontinuity of loss.
To further improve the accuracy of our method on objects smaller than 10 pixels, we introduce a novel RSDet++ which is consists of a point-based anchor-free rotated object detector and a modulated rotation loss.
Extensive experiments demonstrate the effectiveness of both RSDet and RSDet++, which achieve competitive results on rotated object detection in the challenging benchmarks DOTA1.0, DOTA1.5, and DOTA2.0.
We hope the proposed method can provide a new perspective for designing algorithms to solve rotated object detection and pay more attention to tiny objects.
The codes and models are available at: https://github.com/yangxue0827/RotationDetection.

% 基于五参数或者八参数的旋转目标检测器都存在由于损失不连续性引起的旋转敏感性误差这个问题。
% 我们的工作通过介绍了一种基于八参数法的one-stage旋转目标检测器以及调制旋转损失来解决这个问题，并命名为RSDet。
% 1）之前的方法大多是基于五参数回归来定义旋转框，除了损失的不连续性外还会存在单位的不一致性，而RSDet使用八点法很好的解决了这个问题
% 2）我们的调制旋转损失在五点法和八点法上都获得了提升，证明本方法是普遍适用的。
%为了进一步证明调制旋转损失的可靠性，我们基于该损失提出了点回归的anchor-free的旋转目标检测框架，该框架对于小目标拥有更好的实验效果。我们把这种方法称为RSDet++。
% 大量的实验结果证明了RSDet和RSDet++的有效性，他们在遥感卫星检测数据集Dota 1.0 Dota1.5和Dota2.0上均取得了有竞争力的结果
% 我们希望提出的调制旋转损失和RSDet++可以提供一个解决密集小旋转目标检测的新方向，相关的代码和模型即将公布在https://github.com/yangxue0827/RotationDetection。

\end{abstract}


% \fi
\IEEEpeerreviewmaketitle

\section{Introduction}
%%   Para1 旋转目标简介及困难
%. 目标检测是计算机视觉中一个十分重要的基础任务，他可以被用于许多下游的应用比较ReID，图像分割，人脸识别等等。
%  对于常见的水平目标检测器来说，要去检测一些密集，旋转，微小的目标是十分有挑战的。
%  对比，非常直观的这个问题可以通过设计旋转目标检测器来解决
%  最近在计算机视觉中，出现了很多需要旋转目标检测器的场景，比如文本，人脸，车牌和遥感图像等等。
%  因此，我们的目的是为了设计出一种高效的旋转目标检测器，配合一种更加合理的损失函数，可以用于各种旋转目标检测场景。

Object detection is one of the fundamental tasks in computer vision, which motivates many downstream vision applications such as re-identification, semantic segmentation, and face recognization.
Usual object detectors employ horizontal bounding boxes to describe the position of predicted objects, where the candidates of the center point, width, and height are sufficient to locate them efficiently. 
However, it is challenging for them to detect those rotated, dense, and tiny objects accurately.
For example, recently many rotated object detection benchmarks such as aerial dataset (DOTA \cite{R18_xia2018dota}, UCAS-AOD \cite{zhu2015orientation}, HRSC2016 \cite{liu2017a}), scene text dataset (ICDAR2015 \cite{karatzas2015icdar}, ICDAR2017 \cite{gomez2017icdar2017})have been published.
Intuitively, the rotated object detection that provides more accurate position descriptions can solve the above problem.
We aim to design a more reasonable loss function for an effective rotated object detector, which provides a rotated object detection baseline.

%% Para2 常见的旋转目标检测器以及存在的缺点
% 常见的旋转目标检测器可以分为两种：五参数回归法和八参数回归法。
% 五参数回归法适用中心点坐标，宽，高以及旋转角度来描述旋转框的位置，但是这种方法存在回归参数不一致和损失不连续两种缺点。
% 回归参数不一致是因为不同的参数拥有不同的单位，他们具有不同的物理含义，平等的对待他们会导致某种程度的信息损失。
% 损失函数的不连续性，是由于框的定义方式和损失函数之间存在矛盾，连续的角度量和不连续的损失函数会导致在角度边界处出现损失跃变。
% 八参数法使用框的坐标来描述框的位置，解决了回归参数的不一致性，但是仍然存在损失函数的不连续性。

\begin{figure}[tb!]
    \begin{center}
        \includegraphics[width=0.5\linewidth]{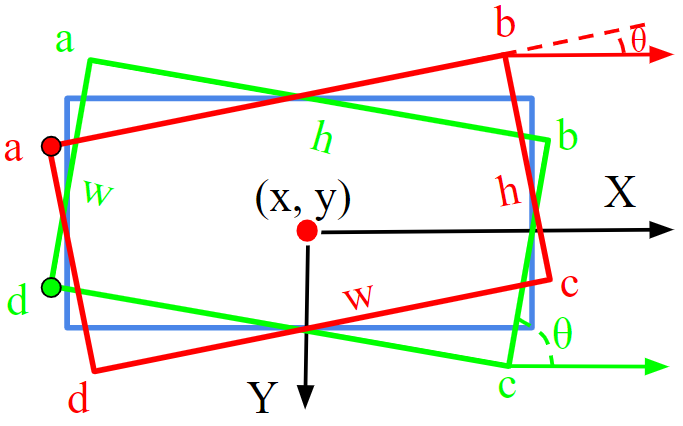}
    \end{center}
    \vspace{-5pt}
    \caption{Loss discontinuity: rectangles in blue, red, and green denote reference box, ground truth, and prediction respectively. 
    Here the reference box is rotated one degree clockwise to get the ground truth and is rotated similarly counterclockwise to obtain the prediction. 
    Then the three boxes are described with five parameters: reference (0, 0, 10, 25, -90$^{\circ}$), ground truth (0, 0, 25, 10, -1$^{\circ}$), and prediction (0, 0, 10, 25, -89$^{\circ}$). Here $\ell_1$ loss is far more than 0.}
    \label{fig:discontinuity_}
    \vspace{-10pt}
\end{figure}

The definition of parameters for rotated object detection has a significant effect on final prediction results, 
and most of the rotated object detectors can be classified as five-parameter or eight-parameter methods based on their parameterization approaches.
Five-parameter methods use coordinates of the central point, width, height, and rotation angle to describe the position and orientation of the rotated bounding box.
However, traditional five-parameter rotated object detection methods have several drawbacks which will influence the final performance,
and we classify them as
: 1) the discontinuity of loss 2) the inconsistency units of regression parameters.
The discontinuity of loss mainly results by the contradiction between the parameterization of the rotated bounding box and the loss function, 
which both neglect the boundary situation during the regression process and lead to a value mutation of loss during training (see Fig.~\ref{fig:discontinuity_}).
Moreover, the parameters, i.e., angle, width, height, and candidates of central point have different measurement units and lead to several relations against the Intersection over Union (IoU). 
These parameters result in inconsistent loss values while previous works add them directly without any specification, which leads to a performance decrease.

Although eight-parameter methods use the coordinates of conners as the regression parameters based on consistent units, 
a similar discontinuity of loss still exists in eight-parameter ones.
The conner order changes with slightly angular disturbance near the boundary position, and the computation of loss can not hold this situation adaptively.

%% Para3 我们的解决方法
%。我们把这种由于损失函数和框的定义导致的损失称为旋转敏感损失，这种损失可以导致网络训练的不平稳。
%   为了解决这个误差，我们提出了调制旋转损失，在这个损失中仔细考虑了旋转过程中存在的边界情况。
%   换句话说，我们在原来的损失后面添加了一个修正项。当旋转位置远离边界位置的时候，这个修正项远远大于真实损失；而当接近边界位置的时候，修正项变得接近真实值。
%   最后，我们只需要对修正项和原损失取最小值就可以获得修正后的损失函数，调制旋转损失的曲线是连续的。

As mentioned before, the discontinuity of loss exists in both five-parameter methods and eight-parameter methods and we summarize this phenomenon as rotation sensitivity error (RSE).
RSE can result in loss mutation when going through the boundary position, and can lead to training instability (see Fig.~\ref{fig:loss_curve}).
In order to address the rotation sensitivity error, we propose a novel modulated rotation loss $\ell_{mr}$ which handles the boundary constraints for rotation carefully, 
leading to a smoother loss curve during the training process. 
In other words, we add a correction term to the original loss and take the minimum value between the primary one and the correction term. 
This correction is particularly greater than $\ell_1$-loss when it does not reach the range boundary of the angle. However, this correction becomes normal when $\ell_1$-loss undergoes a mutation. 
All in all, such correction can be seen as the symmetry of $\ell_1$-loss about the location of the mutation. 
Finally, the curve of $\ell_{mr}$ is continuous after taking the minimum of $\ell_1$-loss and the correction term.

%   para4 方法总述以及如何进一步提升本方法
% 我们的方法是基于Retina-net的，主要是由于一阶段检测器同时满足了对于精度速度的需求。
% 需要注意的是，许多其他的框架比如faster rcnn，yolo等也都可以经过简单的修改应用于我们的工作。
% 近年来，旋转目标检测的数据集越来越多的关注于一些小样本（长或宽小于十个像素的样本），而现有的旋转目标检测器都是基于anchor计算IOU来进行样本匹配的，
%这种基于IOU的匹配方式会随着网络的特征下采样丧失对于小样本的检测能力。
% 这种挑战同样存在于我们的方法中，因为retina-net也是基于anchor进行样本匹配的。

We instantiate our method based on the popular object detector RetinaNet weighting for the performance and speed, termed RSDet.
It's easy to use other detectors (YOLO, Faster RCNN, SSD, CenterNet, and FCOS) as the backbone with just a few modifications.
Recently, the rotated object detection benchmarks focus more on the tiny samples smaller than 10 pixels,
and existing rotated object detectors are all anchor-based methods that have limited performance on these scenes.
Anchor-based methods employ an FPN-architecture for multi-scale objects and use the lower layers to generate anchors for saving the computation cost.
However, the ignorance of top feature maps will lead to performance damage since anchor-based methods map the samples by computing the Intersection over Union (IOU) between predicted and ground-truth boxes.
The tiny ground truth will have a small IOU value with a large anchor, and no suitable anchor matched for the ground truth can be found.
A similar phenomenon can also be found in our RSDet that some tiny objects will be ignored.
%The drawbacks caused by IOU-based mapping approach can also be found in our method since that RetinaNet is also an anchor-based method.

% Para5 RSDet++
% 为了进一步提高本方法对于小样本目标检测的精度，我们提出了基于调制旋转损失的anchor-free 旋转目标检测器，并把他命名为RSDet++。
% 该检测器也是基于八参数回归方法的，避免了五参数法会导致的参数不一致性。
% RSDet++从一个点来回归出旋转目标的四个顶点目标，并且通过调制旋转损失避免了旋转敏感性误差。
% 值得注意的是RSDet++通过点与目标框的位置关系来进行匹配，从而避免了由基于anchor的IOU匹配带来的误差，提高了对于小目标样本的检测能力。

For further improving the performance on tiny samples, a point-based rotated object detection method based on modulated rotation loss is proposed, termed RSDet++.
RSDet++ uses eight parameters to describe the bounding boxes of objects, which are regressed from a center-point instead of the anchor.
We conduct the sample mapping process among ground truth and predictions in RSDet++ based on the distance among center point and ground-truth box, 
which gets rid of the errors from the IoU-based mapping process and improves the detection performance of tiny samples.

\begin{figure}
\centering
\subfigure[RetinaNet-H (baseline)]
{\includegraphics[width=3.5cm]{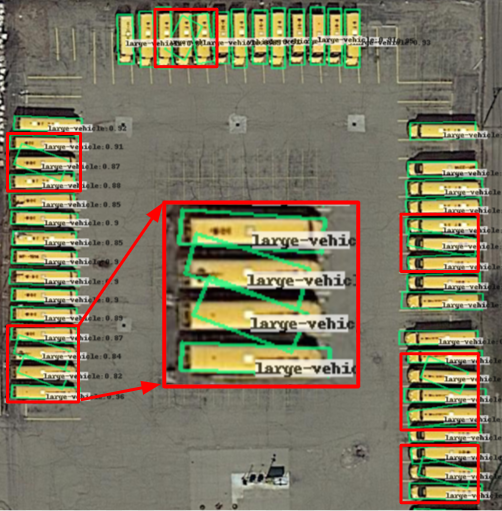}
\label{fig:R_Qloss_vis1}} 
\subfigure[The proposed RSDet]
{\includegraphics[width=3.5cm]{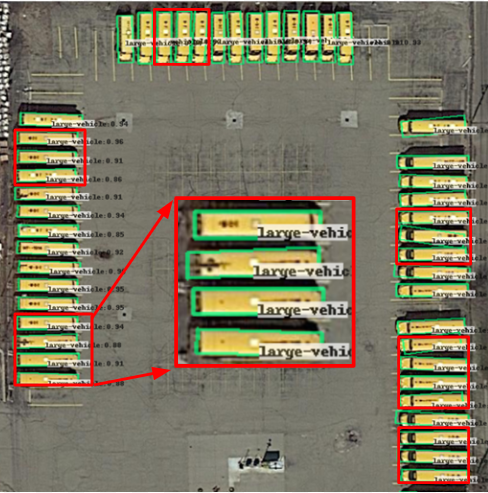}
\label{fig:R_Qloss_vis2}}
\\
\subfigure[RetinaNet-H (baseline)]
{\includegraphics[width=3.5cm]{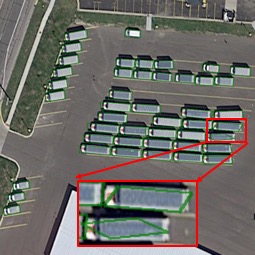}
\label{fig:R_Qloss_vis3}} 
\subfigure[The proposed RSDet++]
{\includegraphics[width=3.5cm]{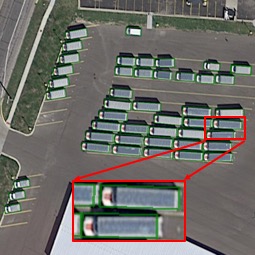}
\label{fig:R_Qloss_vis4}}
\vspace{-5pt}
\caption{Detection results before and after solving the RSE problem with RSDet and RSDet++. 
The red rectangles in (a) and (c) represent failed examples due to the discontinuity of loss.} %图片标题
\vspace{-10pt}
\label{fig:R_Qloss_vis}
\end{figure}

% Para6 contribution
% 1) 我们介绍了常见的五参数回归法和八参数回归法中都会存在的损失函数的不连续性，并将这种性质总结为旋转敏感性误差
% 2) 为了解决旋转敏感性误差，我们提出了调制旋转损失。该损失适用于五参数回归法和八参数回归法。
% 3) 为了进一步提高对于小目标的检测能力，我们基于点回归的框匹配方式和调制旋转损失提出RSDet++，取得了更快和更好的性能

%RSDet and RSDet++ show competitive performance on the DOTA1.0, DOTA1.5, and DOTA2.0 benchmark, and our techniques are all orthogonal to existing methods. \textbf{The contributions of this paper are:}
The preliminary content of this paper has partially appeared in AAAI 2021~\cite{DBLP:conf/aaai/Qian0PYG21}
\footnote{To obtain a more thorough analysis and comprehensive results on tiny objects, the conference version has been significantly extended and improved in the journal version, especially in the following aspects:
i) we point out that anchor-based rotated object detection methods show poor performance on tiny objects, and propose a novel anchor-free RSDet++ to solve the problem;
ii) we use modulated rotation loss in RSDet++, which reaches consistent improvement in anchor-based method;
iii) we use more challenging datasets to verify the performance on tiny objects, such as DOTA1.5 and DOTA2.0, which contain more tiny objects.
}. 
The overall contribution of this extended journal version can be summarized as:
%\begin{enumerate}
   % \item
   
   i) We introduce the discontinuity of loss which exists in both five-parameter and eight-parameter rotated object detection methods, and formally formulate this phenomenon as rotation sensitivity error (RSE).
    
  %  \item
  
    ii) For alleviating the RSE in the widely used five-parameter and eight-parameter systems, we devise a special treatment to ensure the loss continuity. We term the new loss as modulated rotation loss $\ell_{mr}$.

 %   \item

    iii) To further improve the performance of the rotated object detection methods on tiny samples, we propose RSDet++ based on the point-based bounding box regression method and modulated rotation loss.

\section{Related Work}
%在这篇论文中，我们主要讨论旋转目标检测中边界处存在的不连续性以及对于微小目标的漏检情况。
%要想弄清楚这些问题，首先就得搞明白，在水平目标检测中这些问题是存不存在，以及是如何从高水平目标过渡到倾斜目标的
We emphasize two problems of rotated object detection in this paper: the discontinuity of loss and the ignorance of tiny objects.
We analyze whether these problems existing in horizontal object detection methods and why they occur in rotated object detections.
Finally, some methods that focus on a similar problem with us are introduced.

%水平目标检测器
%改到这里了
\paragraph{Horizontal Object Detectors}
Visual object detection has been a hot topic over the decades. Since the seminal work R-CNN \cite{girshick2014rich}, there have been a series of improvements including Fast RCNN \cite{girshick2015fast}, Faster RCNN \cite{R16_Ren2015Faster}, and R-FCN \cite{R15_dai2016r}, which fall into the category of the two-stage methods. 
On the other hand, people develop single-stage approaches which are more efficient than the two-stage methods. 
Examples include Overfeat \cite{sermanet2013overfeat}, YOLO \cite{redmon2016you}, and SSD \cite{R11_liu2016ssd}. 
In particular, SSD \cite{R11_liu2016ssd} combines the advantages of Faster RCNN and YOLO to achieve the trade-off between speed and accuracy. 
Subsequently, multi-scale feature fusion techniques are widely adopted in both single-stage methods and two-stage ones, such as FPN \cite{lin2017feature}, RetinaNet~\cite{R25_Lin2017Focal}, and DSSD~\cite{fu2017dssd}. 
Recently, researchers put forward many cascaded or refined detectors. For example, Cascade RCNN \cite{cai2018cascade}, HTC \cite{chen2019hybrid}, and FSCascade \cite{zhang2018single} perform multiple classifications and regressions in the second stage, leading to notable accuracy improvements in both localization and classification. 
Besides, the anchor free methods have become a new research focus, including FCOS~\cite{R14_Tian2019FCOS}, FoveaBox~\cite{kong2019foveabox}, and RepPoints~\cite{Yang_2019_ICCV}. 
People simplify the structures of these detectors by discarding anchors, so anchor-free methods have opened up a new direction for object detection.

However, the above detectors only generate horizontal bounding boxes, which limit their applicability in many real-world scenarios. 
The objects in scene texts or aerial images tend to be densely arranged and have large aspect ratios, which requires more accurate localization. 
Therefore, rotated object detection has become a prominent direction in recent studies~\cite{Yang_2019_ICCV}.
Moreover, the aforementioned horizontal object detection methods \cite{girshick2014rich, redmon2016you, R11_liu2016ssd, R25_Lin2017Focal} use four parameters (coordinates of the center point, width, and height) to describe the position of an object, which represent different physical meanings with no boundary condition needs to be considered.
For improving the performance on small objects, horizontal object detection methods usually use the FPN as backbone \cite{R15_dai2016r, lin2017feature}, and a cascaded network architecture \cite{cai2018cascade, chen2019hybrid, zhang2018single} is usual.

%旋转目标检测器

\paragraph{Rotated Object Detectors}
Rotated object detection usually uses an extra angle parameter to describe the orientation of an object or use the coordinates of four vertexes which embedding the orientation information,
and rotated bounding boxes can provide a more accurate position description than horizontal bounding boxes.
Rotated object detection has been widely used in natural scene text \cite{R22_Jiang2017R2CNN, R30_ma2018arbitrary}, aerial imagery \cite{fu2018ship,yang2018object,yang2019building}, etc. 
People introduce many excellent detectors for scene text. 
For example, RRPN \cite{R30_ma2018arbitrary} uses rotating anchors to improve the qualities of region proposals. R$^2$CNN \cite{R22_Jiang2017R2CNN} is a multi-tasking text detector that simultaneously detects rotated and horizontal bounding boxes. 
In TextBoxes++~\cite{R31_liao2018textboxes++}, a long convolution kernel is used to accommodate the slenderness of the text with the number of proposals increasing. 
% EAST \cite{Zhou2017EAST} proposes a simple yet powerful pipeline that yields fast and accurate text detection in natural scenes. 

Moreover, object detection in aerial images is more challenging for the complex backgrounds, dense arrangements, and a high proportion of small objects. 
Many scholars have also applied general object detection algorithms to aerial images, and many robust rotated detectors have emerged. 
For example, ICN \cite{R27_azimi2018towards} combines various modules such as image pyramid, feature pyramid network, and deformable inception sub-networks, and it achieves satisfactory performances on the DOTA benchmark. 
RoI Transformer \cite{R29_ding2018learning} extracts rotation-invariant features for boosting subsequent classification and regression. 
SCRDet \cite{R28_Yang2018SCRDet} proposes an IoU-smooth $\ell_1$ loss to solve the sudden loss change caused by the angular periodicity so that it can better handle small, cluttered, and rotated objects. 
R$^3$Det \cite{R20_Yang2019R3Det} proposes an end-to-end refined single-stage rotated object detector for fast and accurate object localization by solving the feature misalignment problem.

\begin{figure}[tb!]
\centering
\subfigure[Width is longer than height. ]{\includegraphics[height=3cm,width=4cm]{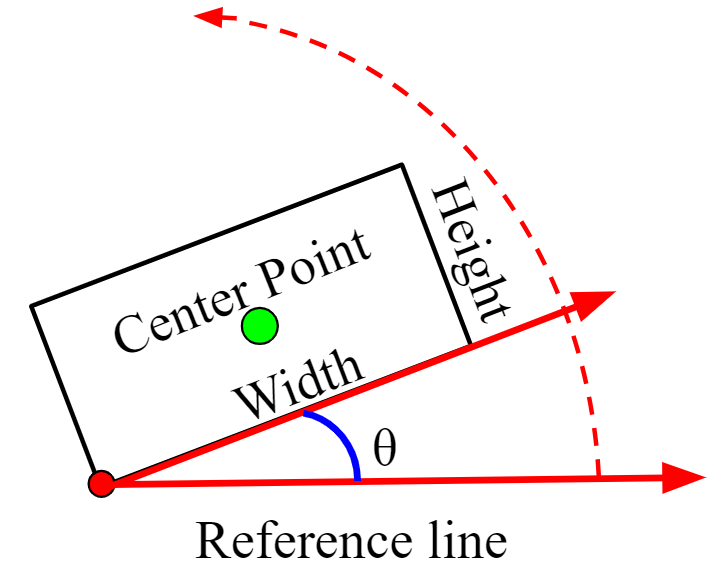}} 
\label{fig:3_1}
\subfigure[Height is longer than width.]{\includegraphics[height=3cm,width=4cm]{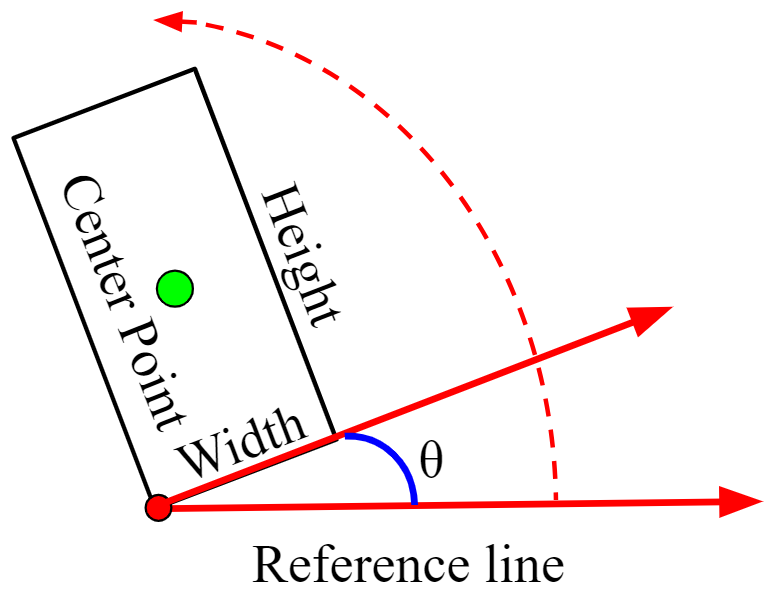}}
\label{fig:3_2}
\vspace{-5pt}
\caption{The five-parameter definition in OpenCV exchanges the width and the height in the boundary condition for rotation. 
The angle parameter $\theta$ ranges from -90 degree to 0 degree, but it should be distinguished from another definition \cite{R18_xia2018dota}, with 180 degree angular range, 
whose $\theta$ is determined by long side of rectangle and x-axis.} %图片标题
\label{fig:define}
\vspace{-10pt}
\end{figure}

\paragraph{The ignored boundary condition}
All the above mentioned rotated object detectors \cite{R22_Jiang2017R2CNN, R29_ding2018learning, R20_Yang2019R3Det} do not take the inherent discontinuity of loss into account, 
which can damage learning stability and final detection performances in our experiments. 
However, no existing studies have addressed this fundamental problem that motivates our work.
Moreover, we notice some works deal with a similar problem with us.
SCRDet \cite{R28_Yang2018SCRDet} combines the IOU-loss and smooth L1 loss, and the IoU-smooth L1 loss partly circumvents the need for differentiable rotating IoU loss.
Gliding vertex \cite{2020Gliding} and CenterMap \cite{wang2020learning} use quadrilateral and mask to accurately describe oriented objects, respectively.
CLS \cite{yang2020arbitrary} transforms angular prediction from a regression problem to a classification one.
We operate a comparison between our RSDet and the above methods, which shows that our method outperforms them.

%旋转目标检测器中常见的问题

%\section{Related Work}

\section{Methodology}
%前言部分
%首先，我们重新介绍了旋转目标检测中常见的两种目标框定义方式并且介绍了他们各自存在的缺陷在section3.1
%第二，我们介绍了调制旋转损失的概括性公示，并且详细介绍了他们在不同的旋转框定义中的具体表达方式。
%第三，我们介绍了基于retinanet和八点调制旋转损失合并的网络结构，并把它称为RSdet
%第四，section3.4节中介绍了anchor-based方法普遍容易忽略小目标这个问题，以及提出了基于点回归的RSDet++版本
Firstly, we reformulate popular rotated object detection methods can be classified as five-parameter methods and eight-parameter methods, 
and summarize the discontinuity of loss as the rotation sensitivity error which commonly exists in previous methods in section3.1.
Secondly, we introduce a modulated rotation loss to alleviate the rotation sensitivity loss and propose the RSDet based on modulated rotation loss in section3.2.
Finally, we propose RSDet++, which employs a point-based bounding box regression method to improve the detection performance on tiny objects.

\subsection{Parameterization of Rotated Bounding Box}

\subsubsection{Five-parameter Methods}
%第一段介绍五参数法
%常见的五参数定义法主要是延伸了opencv中的定义，如图三a）中所示
Given a rotated object, and we aim to describe its position and orientation by five parameters:
the coordinates $(x, y)$ of the center point, the width $w$, and the height $h$ for the positional description;
the rotation angle $\theta$ for the orientation description.
This kind of definition is in line with that in OpenCV as shown in Fig.~\ref{fig:define}, while the width $w$ and the height $h$ are related to the angle $\theta$ :
a) define the reference line along the horizontal direction that locates the vertex with the smallest vertical coordinate. 
b) rotate the reference line counterclockwise, the first rectangular side that be got to by the reference line is width $w$ regardless of its length compared with the other side -- height $h$. 
c) the central point coordinate is $(x, y)$, and the rotation angle is $\theta$.

\begin{figure}[!tb]
\centering
\subfigure[Relation between center point and IoU]{\includegraphics[width=2.6cm,height=1.8cm]{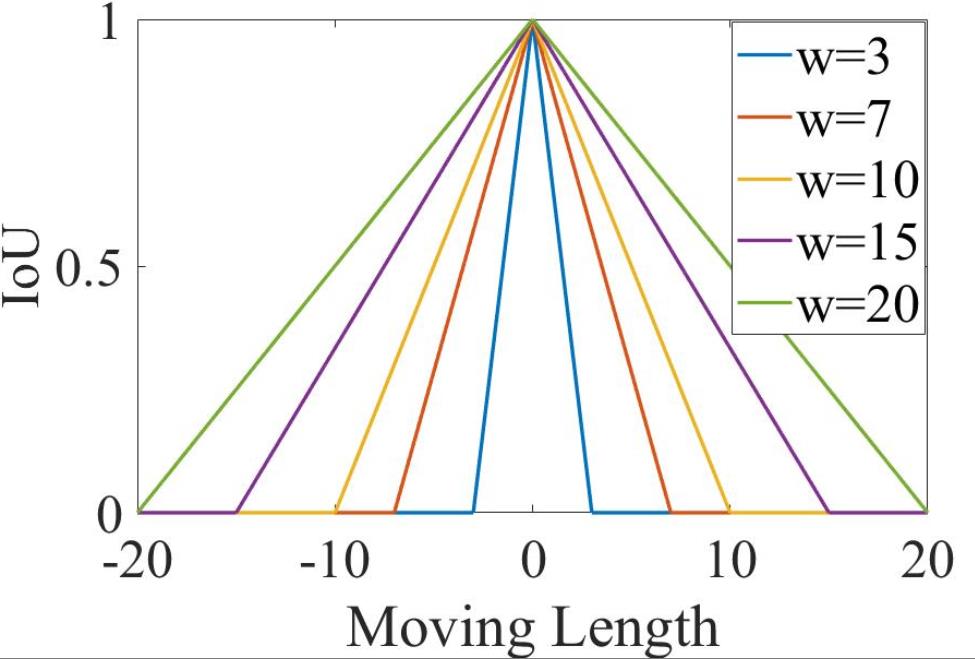}
\label{fig:2_1}} 
\subfigure[Relation between width and IoU]{\includegraphics[width=2.6cm,height=1.8cm]{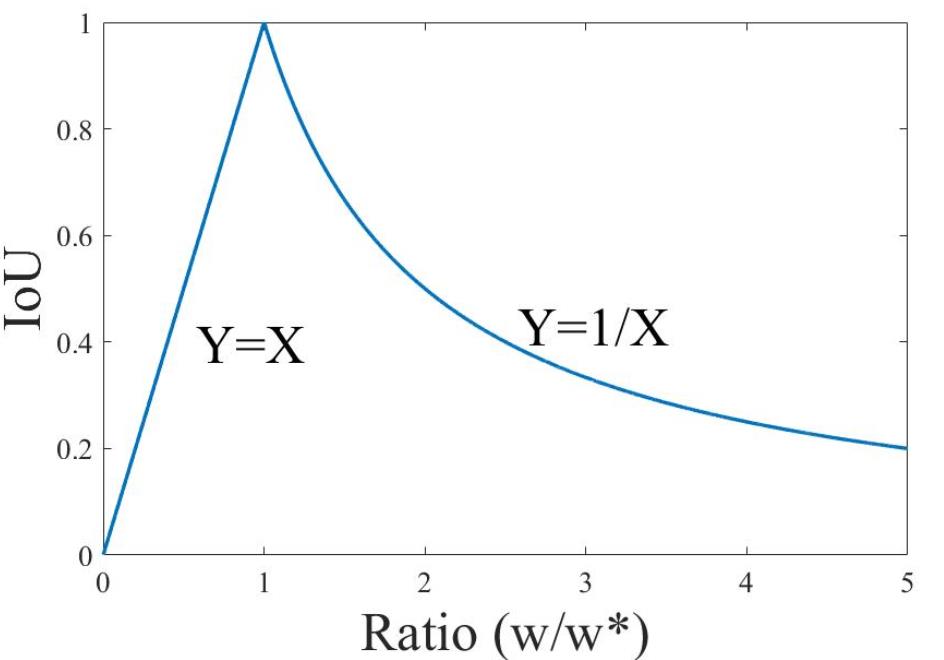}
\label{fig:2_2}}
\subfigure[Relation between angle and IoU]{\includegraphics[width=2.6cm,height=1.8cm]{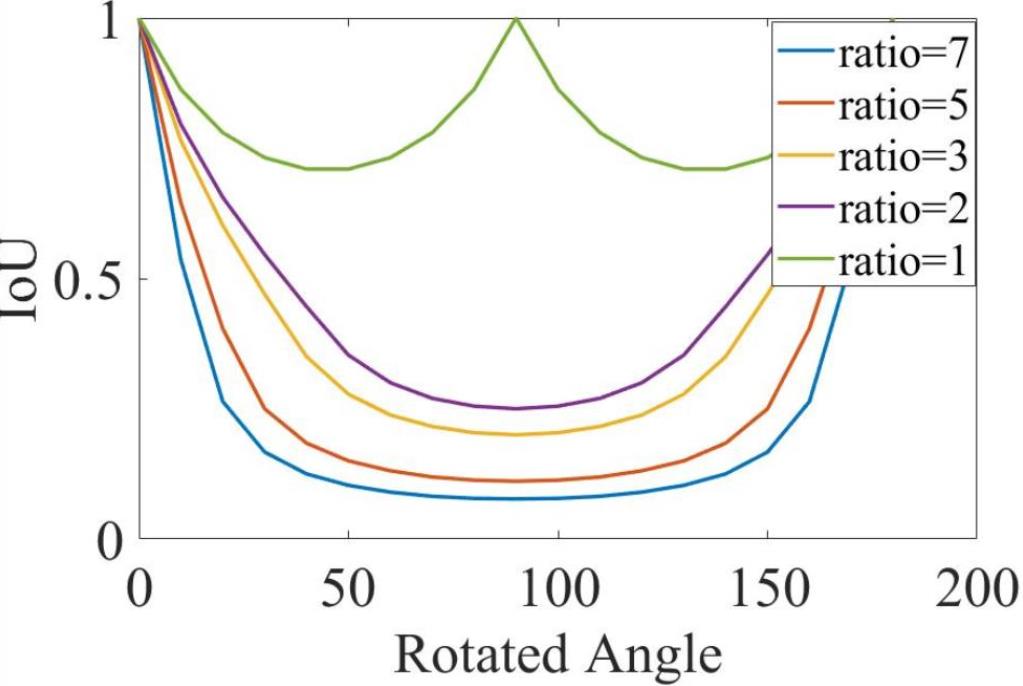}
\label{fig:2_3}} 
%\subfigure[Comparison of three %relations]{\includegraphics[width=4cm]{figure/fig6_IoU.jpg}
%\label{fig:2_4}}
\caption{Inconsistency in five-parameter regression model. The relationship between height and IoU is similar to relationship between width and IoU.}
\label{fig:inconsistency}
\vspace{-10pt}
\end{figure}

Although some previous rotated object detection methods such as R$^2$CNN \cite{R22_Jiang2017R2CNN}, 
RRPN~\cite{R30_ma2018arbitrary}, ICN \cite{R27_azimi2018towards}, and FFA \cite{fu2020rotation} all employ the five parameter parameterization and achieve great improvements when comparing with traditional methods.
There also exist some shortcomings which have negative impacts on detection performance:
%We find that the training process of these methods are unstable, and some reasons are classified as following:
%尽管一些经典的选择目标检测器R2CNN，RRPN，ICN，FFA等等都使用了这样的五参数法取得了不错的实验效果。
%但是我们发现现有的方法在训练过程中并不是特别的稳定，我们分析归纳了下面两点原因。

%1）回归参数的不一致性
The inconsistent of parameter units: 
different measurement units of five parameters lead to inconsistent regression processes, 
and the impact of such artifacts has rarely been studied in the literature. 
We analyze the relationships among all the parameters and IoU empirically in Fig.~\ref{fig:inconsistency}.
Specifically, the relationship between IoU and width (height) behaves like a combination of a linear function and an inverse proportion function, as illustrated in Fig.~\ref{fig:2_1}. 
The relationship between the central point and IoU is a symmetric linear function, as illustrated in Fig.~\ref{fig:2_2}. 
Completely different from other parameters, the relationship between the angle parameter and IoU is a multiple polynomial function (see Fig.~\ref{fig:2_3}). 
Such regression inconsistency is highly likely to deteriorate the training convergence and the detection performance. 
Note that we use IoU as the standard measurement since that the final detection performance depends on whether IoU between the prediction and ground truth is high enough.

%2）损失函数的不连续性
The discontinuity of loss: the angle parameter is the main reason for the loss discontinuity. 
To obtain the predicted box that coincides with the ground truth box, the horizontal reference box rotates counterclockwise, as shown in Fig.~\ref{fig:ASE}. 
In this figure, the coordinates of the predicted box are transformed from those of \textcolor{blue}{the blue reference box} $(0,0,100,25,-90^\circ)$ to $(0,0,100,25,-100^\circ)$ in the normal coordinate system. 
However, the angle of the predicted box is out of the defined range, and the actual coordinates of \textcolor{green}{the green ground truth} box are $(0,0,25,100,-10^\circ)$. 
Although the rotation process is physically smooth, the loss becomes large since the discontinuity of loss. 
To avoid such a loss fluctuation, the reference box need to be rotated clockwise to obtain the gray box $(0,0,100,25,-10^\circ)$ in Fig.~\ref{fig:ASE1} (step 1), then width and height of the gray box will be scaled to obtain \textcolor{red}{the red predicted box} $(0,0,25,100,-10^\circ)$ (step 2). 
At this time, although the loss value is close to zero, the detector experiences a complex regression. 
The above process requires relatively high robustness, which increases the training difficulty. 
More importantly, More importantly, an explicit and specific way for achieving a smooth regression process is needed.

\subsubsection{Eight-parameter Methods.}
%第一段介绍八点法
%基于八参数法的旋转目标检测器主要是通过对四个顶点位置的预测来描述旋转矩形的位置。
%通常来讲，这四个顶点要求按照某种固定的顺序进行排列，描述如下：
%starting from the lower left corner, four clockwise vertices (a, b, c, d) of the rotated bounding box are used to describe its location, as shown in Fig.~\ref{fig:discontinuity_}. 
%In fact, such a parameterization protocol is convenient for quadrilateral description, which is friendly in more complex application scenarios.
%尽管八点法在之前的水平目标检测器中已经被提出来了，但是旋转目标检测器中却很少提及。
Instead of the above five-parameter methods, the eight-parameter method employs the coordinates of vertexes to describe the position and orientation of rotated objects.
To be consistent with the description, we order the four vertexes as following:
a) starting from the lower-left corner a
b) four clockwise vertices (a, b, c, d) of the rotated bounding box are used to describe its location, as shown in Fig.~\ref{fig:discontinuity_}. 
In fact, such a parameterization protocol is convenient for quadrilateral description, which is friendly in more complex application scenarios.
%It is worth mentioning that although the eight-parameter method has been proposed in some horizontal object detectors, it is rarely applied in the rotated object detectors.

\begin{figure}[!tb]
\centering
\subfigure[5-parameter regression w/ two steps in boundary condition]{
            \label{fig:ASE1}
            \includegraphics[width=3.5cm,height=3.5cm]{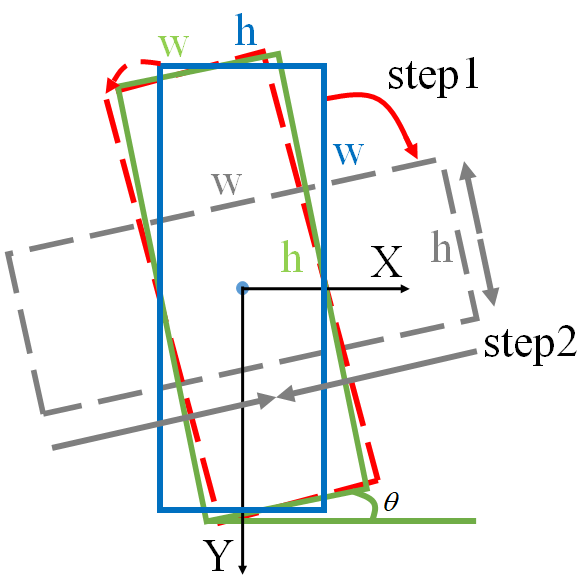}} 
\subfigure[Eight-parameter regression procedure]{
            \label{fig:ASE2}
            \includegraphics[width=3.5cm,height=3.5cm]{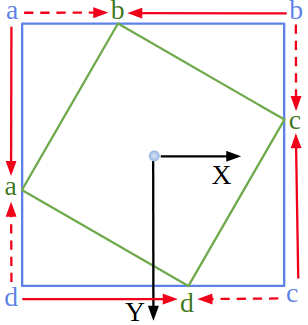}}
\vspace{-5pt}
\caption{Boundary discontinuity analysis of five-parameter regression and eight-parameter regression. The red solid arrow indicates the actual regression process, and the red dotted arrow indicates the ideal regression process.} %图片标题
\label{fig:ASE}
\vspace{-10pt}
\end{figure}

%相比较于5参数法旋转目标检测器，八参数法由于所有的参数都表示的点的位置关系拥有相同的单位。
%但是我们分析发现，损失函数的不连续性在八参数法中也同样存在
Although the parameter units of eight-parameter methods are consistent, 
we find that the loss discontinuity also exists.
Supposing that a ground truth box can be described with the corner sequence $a\rightarrow b\rightarrow c\rightarrow d$ (see red box in Fig.~\ref{fig:discontinuity_}). 
However, the corner sequence becomes $d\rightarrow a\rightarrow b\rightarrow c$ (see green box in Fig.~\ref{fig:discontinuity_}) when the ground truth box undergoes a slightly angle change.
Therefore, consider the situation of an eight-parameter regression in the boundary case, as shown in Fig.~\ref{fig:ASE2}. 
The actual regression process from the blue reference box to the green ground truth box is $\{(\textcolor{blue}{a}\rightarrow \textcolor{green}{a}),(\textcolor{blue}{b}\rightarrow \textcolor{green}{b}),(\textcolor{blue}{c}\rightarrow \textcolor{green}{c}),(\textcolor{blue}{d}\rightarrow \textcolor{green}{d})\}$, 
but apparently the ideal regression process should be $\{(\textcolor{blue}{a}\rightarrow \textcolor{green}{b}),(\textcolor{blue}{b}\rightarrow \textcolor{green}{c}),(\textcolor{blue}{c}\rightarrow \textcolor{green}{d}),(\textcolor{blue}{d}\rightarrow \textcolor{green}{a})\}$. 
This situation also causes the regression process more difficult and unstable.

\subsubsection{Rotation Sensitivity Error.}
%基于上述对于五参数法和八参数法的介绍，我们不难发现，损失的不连续性在旋转目标检测中是个常见的问题，我们将他归纳为旋转敏感性误差。
%现有的旋转框的定义方式基本沿用水平框的定义方式，这就导致他们与旋转目标检测本质上是不适配的，我们对旋转敏感性误差的产生原因进行了一些分析：
%1）水平框检测定义的中心点和宽高，分别代表不同的物理意义并且相互之间是没有依赖性的。
%而旋转目标检测由于角度的原因，无论是五参数法中的角度和别的参数值，还是八参数法中的四个点，他们的定义都是相互依赖的。
%这种依赖关系导致在某些边界位置处，参数之间没法对应上，所以在计算损失的时候就会出现损失的跃变.
Based on the above analysis of five-parameter and eight-parameter methods, 
we conclude the discontinuity of loss as the rotation sensitivity error in rotated object detection.
The existing rotated bounding box definition method is modified from 
previous horizontal object detection, and it is unsuitable for rotated object detection.

The coordinates of the object center $(x, y)$, the width $w$, and the height $h$ for horizontal object position description are independent and have exclusive physical meaning.
However, the definition of the width $w$ and the height $h$ are dependent on the angle $\theta$ in five-parameter methods, 
while the coordinates of vertexes in the eight-parameter mode share the same physical meaning and confuse the loss computation process.
When computing for the rotated object detection losses, 
the parameters for the predicted bounding box and the ground-truth box may be misaligned near the boundary condition for the ASE, which leads to the discontinuity of loss.

\subsection{Our Proposed Modulated loss}
%要解决这个问题，有两种解决方法，1）提出一种新的旋转目标定义方式可以避免旋转敏感误差 2）依据现有的方法，在损失中进行调节，避免边界处的损失跃变。
%我们认为第二种方案是更有价值的，因为现有的大量的检测框架是基于之前定义方式搭建的，而通过提出一种新的损失可以很大程度降低研发成本。
%基于上述分析，我们提出了一种调制旋转损失来解决旋转敏感性误差。
Two kinds of approaches are employed to solve the rotation sensitivity error (RSE):
1) a novel definition for the rotated bounding box which can avoid the rotation sensitivity error;
2) correcting the boundary condition through loss function based on existing parameterization methods.
Considering that there are many existing rotated detection methods based on five-parameter or eight-parameter parameterization,
we propose modulated rotation loss $\ell_{mr}$ to solve RSE in this section.
The pseudo equation of $\ell_{mr}$ can be described as:
\begin{equation}
    \begin{array}{l}
        \ell_{mr}=\min\left\{\begin{array}{l}
        \ell_{1}(P_r,\; T) 
        \\           
        \ell_{1}(P_m,\;T),
        \end{array}\right.
    \end{array}
\end{equation}
where $P_r$ and $P_m$ represent the original predicted box and the modulated predicted box, $T$ represents the ground-truth or label, and $\ell_{1}$ represents for the least absolute error.

\subsubsection{Five-parameter Modulated Rotation Loss}
\begin{figure}[!tb]
\centering
\subfigure[Discontinuous $\ell_1$-loss]{\includegraphics[width=4cm]{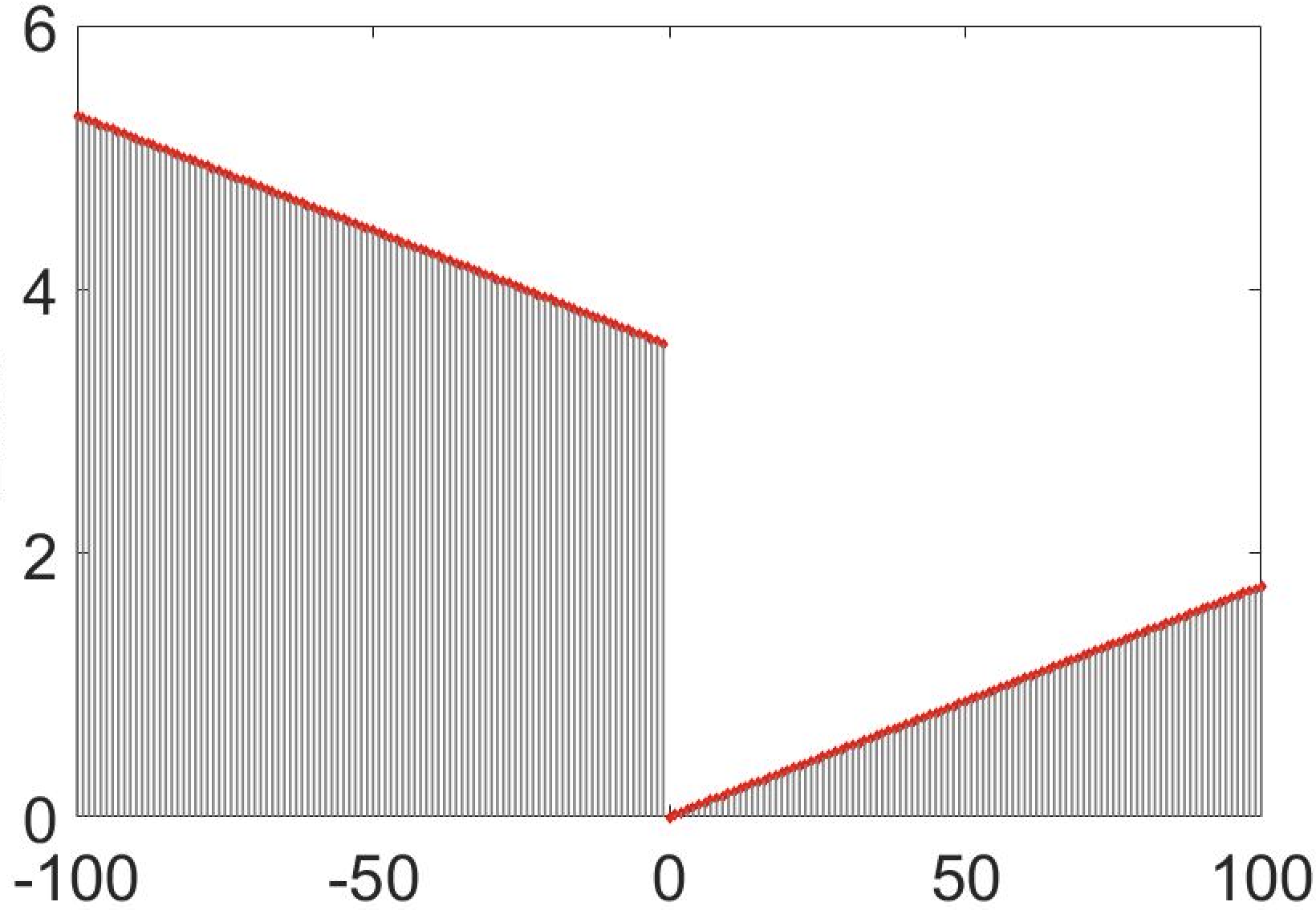}
\label{fig:fig4a}} 
\subfigure[Continuous $\ell^{5p}_{mr}$]{\includegraphics[width=4cm]{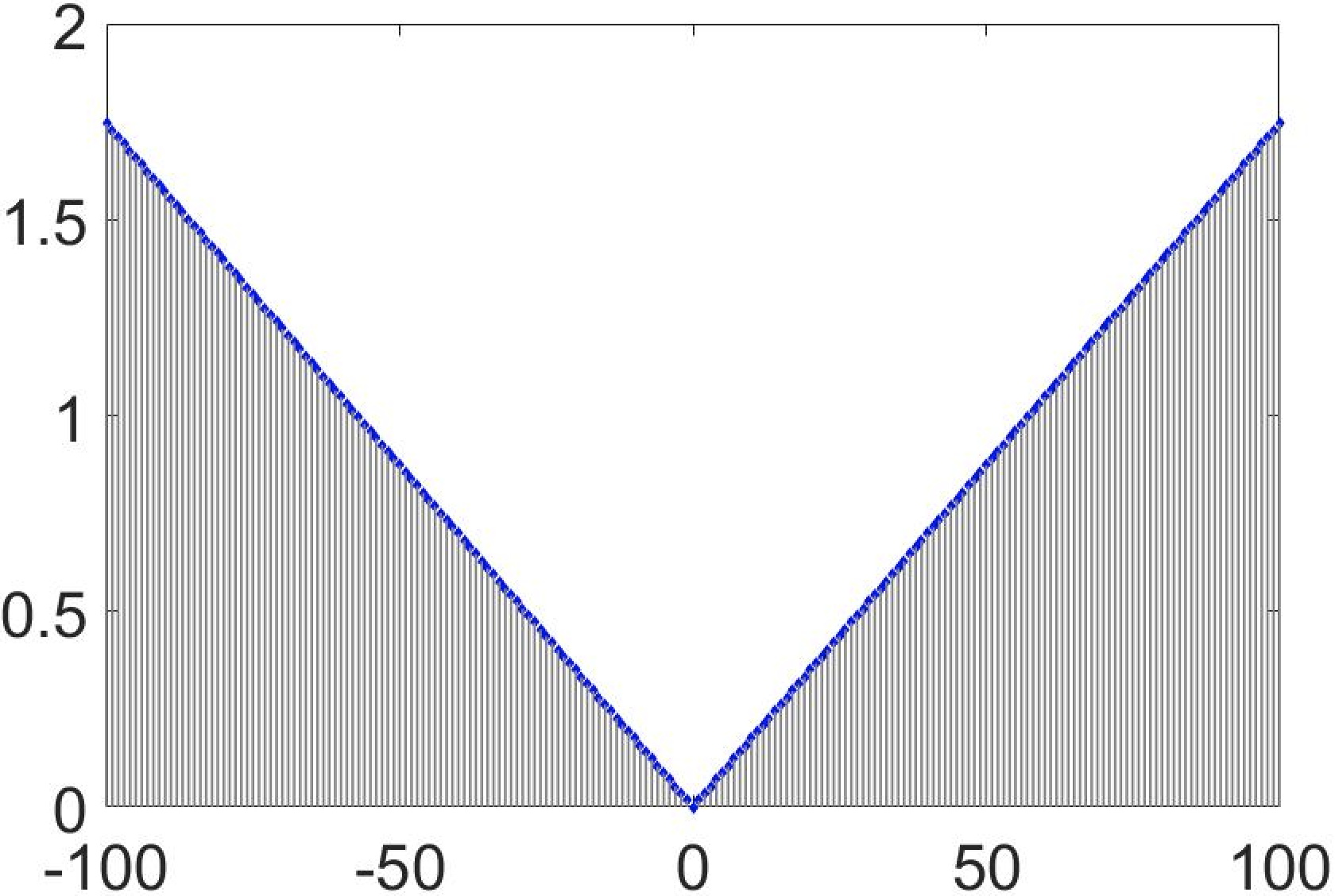}
\label{fig:fig4b}}
\vspace{-5pt}
\caption{Comparison of loss continuity between two kinds of loss functions.} %图片标题
\label{fig:discontinuity}
\vspace{-10pt}
\end{figure}

As mentioned above, RSE only occurs in the boundary cases (see Fig.~\ref{fig:fig4a}). In this paper, we devise the following boundary constraints to modulate the loss as termed by modulated rotation loss $ \ell^{5p}_{mr}$:
\begin{equation}
    \begin{aligned}
        \ell_{cp}=|x_{1}-x_{2}|+|y_{1}-y_{2}|,
    \end{aligned}
\end{equation}
\begin{equation}
    \begin{array}{l}
        \ell^{5p}_{mr}=\min\left\{\begin{array}{l}
         \ell_{cp}+|w_{1}-w_{2}|+|h_{1}-h_{2}| 
        +|\theta_{1}-\theta_{2}|
        \\           
        \ell_{cp}+|w_{1}-h_{2}|+|h_{1}-w_{2}| 
        \\ \quad \quad \quad +|90-|\theta_{1}-\theta_{2}||,
        \end{array}\right. 
    \end{array}
\end{equation}
where $\ell_{cp}$ is the central point loss. The first item in $\ell_{mr}$ is $\ell_1$-loss. The second item is a correction used to make the loss continuous by eliminating the angular periodicity and the exchangeability of height and width. This correction is particularly larger than $\ell_1$-loss when it does not reach the range boundary of the angle parameter. 
However, this correction becomes normal when $\ell_1$-loss changes abruptly. 
In other words, such correction can be seen as the symmetry of $\ell_1$-loss about the location of the mutation. 
Finally, $\ell_{mr}$ takes the minimum of $\ell_1$-loss and the correction term. 
The curve of $\ell_{mr}$ is continuous, as sketched in Fig.~\ref{fig:fig4b}.

%8/12/2021 9:44 review 

In practice, the employment of relative values of bounding box regression can avoid the errors caused by objects on different scales. 
Therefore, the $\ell_{mr}$ in this paper is:

\begin{equation}
    \begin{aligned}
    \nabla \ell_{cp}=|t_{x1}-t_{x2}|+|t_{y1}-t_{y2}|,
    \end{aligned}
\end{equation}
    \begin{equation}
        \begin{aligned}
        \ell^{5p}_{mr}=\min\left\{\begin{array}{l}
        \nabla \ell_{cp}+|t_{w1}-t_{w2}|
        +|t_{h1}-t_{h2}|+|t_{\theta 1}-t_{\theta 2}|
        \\%[1mm]
        \nabla \ell_{cp}+|t_{w1}-t_{h2}-\log(r)| \\%[1mm]
        +|t_{h1}-t_{w2}+\log(r)| +||t_{\theta 1}-t_{\theta 2}| -\frac{\pi}{2}|,
        \end{array}\right. 
        \end{aligned}
        \label{eq:l_5pmr}
\end{equation}
    where
    $t_{x}=(x-x_{a})/w_{a},  t_{y}=(y-y_{a})/h_{a}$, $t_{w}=\log(w/w_{a}),  t_{h}=\log(h/h_{a}), 
    r=\frac{h}{w},  t_{\theta}=\frac{\theta \pi}{180}$.
    Here the measurement unit of the angle parameter is radian, $r$ represents the aspect ratio. $x$ and $x_{a}$ are respectively the predicted box and the anchor box (likewise for $y$, $w$, $h$, $\theta$).

\subsubsection{Eight-parameter Modulated Rotation Loss}
The eight-parameter regression method means the prediction is a quadrilateral. 
Quadrilateral regression needs to sort the four corner points in advance, or else we will reach a huge loss value even if the position prediction is correct. 
For solving the ambiguity of point order, we propose an algorithm to get the sequence of four points (detailed implementation as shown in algorithm \ref{alg:points_sequence}). 
Note that this algorithm is only for convex quadrilaterals and takes clockwise output as an example.
The loss discontinuity phenomenon also exists in the eight-parameter regression method,
and we devise the eight-parameter version of our modulated rotation loss which consists of three components: i) move the four vertices of the predicted box clockwise by one place; ii) keep the order of the vertices of the predicted box unchanged; iii) move the four vertices of the predicted box counterclockwise by one place; iv) take the minimum value in the above three cases. Therefore, $\ell^{8p}_{mr}$ can be expressed as:
    \begin{equation}
%\begin{footnotesize}
    \ell^{8p}_{mr}=\min\left\{\begin{split}
    &\sum_{i=0}^{3}\left( \frac{|x_{(i+3)\%4}-x^*_{i}|}{w_a}+\frac{|y_{(i+3)\%4}-y^*_{i}|}{h_a}\right)
    \\
    &\sum_{i=0}^{3}\left(\frac{|x_{i}-x^*_{i}|}{w_a}+\frac{|y_{i}-y^*_{i}|}{h_a}\right)
    \\
    &\sum_{i=0}^{3}\left(\frac{|x_{(i+1)\%4}-x^*_{i}|}{w_a}+\frac{|y_{(i+1)\%4}-y^*_{i}|}{h_a}\right)
    \end{split}\right.
    \label{eq:l_8pmr}
%\end{footnotesize}
    \end{equation}
where $x_{i}$ and $y_i$ denote the $i$-th vertex coordinates of the predicted box and the reference box. $x^*_i$, $y^*_i$ respectively denote the $i$-th vertex coordinates of the ground truth box and the reference box.

Through the eight-parameter regression and the definition of $\ell^{8p}_{mr}$, the problems of the regression inconsistency and the loss discontinuity in rotation detection are eliminated. Extensive experiments show that our method is more stable for training (see Fig.~\ref{fig:loss_curve}) and outperforms other methods (as shown in Table~\ref{tab:dota_sota}).

\begin{algorithm}[t]
		\caption{Determination of the sequence of quadrilateral corner points.} 
		\label{alg:points_sequence}
		\hspace*{0.02in} {\bf Input:} 
		Four vertex vectors of quadrilateral $\vec{p}_{1},\vec{p}_{2},\vec{p}_{3},\vec{p}_{4}$\\
		\hspace*{0.02in} {\bf Output:} 
		Four vertex vectors arranged clockwise \\ $\vec{p'}_{1},\vec{p'}_{2},\vec{p'}_{3},\vec{p'}_{4}$
		\begin{algorithmic}[1]
			\State $S \leftarrow \{\vec{p}_{1},\vec{p}_{2},\vec{p}_{3},\vec{p}_{4}\}$, $\vec{p'}_{2}=\vec{p'}_{3}=\vec{p'}_{4}=\vec{\textbf{0}}$;
			\State $\vec{p'}_{1} \leftarrow FindLeftmostVertex(S)$;
			\State $S \leftarrow S-\{\vec{p'}_{1}\}$
			
			\For{$\vec{s}_{1} \in S$}
			\State $\vec{s}_{2},\vec{s}_{3} \in S - \{\vec{s}_{1}\}$
			\If{$ CrossProduct(\vec{s}_{1}-\vec{p'}_{1}, \vec{s}_{2}-\vec{p'}_{1}) \times CrossProduct(\vec{s}_{1}-\vec{p'}_{1}, \vec{s}_{3}-\vec{p'}_{1}) < 0$}
			
			\State $\vec{p'}_{3}=\vec{s}_{1}$, $S \leftarrow \{\vec{s}_{2},\vec{s}_{3}\}$;
			\State break;
			\EndIf
			\EndFor
			
			\For{$\vec{s}_{1} \in S$}
			\State $\vec{s}_{1}=S-\{\vec{s}_{1}\}$;
			\If{$CrossProduct(\vec{p'}_{3}-\vec{p'}_{1}, \vec{s}_{1}-\vec{p'}_{1}) > 0$}
			\State $\vec{p'}_{2}=\vec{s}_{1}$, $\vec{p'}_{4}=\vec{s}_{2}$;
			\Else
			\State $\vec{p'}_{2}=\vec{s}_{2}$, $\vec{p'}_{4}=\vec{s}_{1}$;
			\EndIf
			\EndFor
			\State \Return $\vec{p'}_{1},\vec{p'}_{2},\vec{p'}_{3},\vec{p'}_{4}$
		\end{algorithmic}
	\end{algorithm}

\subsubsection{RSDet based on Modulated Rotation loss}
%我们注意到尽管八点法很早就被提出了，但是在旋转目标检测领域并没有相关的代码以及工作。
%因此我们基于retinanet修改了八参数版本的，并且配合modulated rotation loss提出了RSDet。
%除此之外我们还提出了基于5参数法的RSdet，这两种版本的RSdet在配合modulated rotation loss之后均可以获得一致的提高。
The proposed RSDet is an effective and unified framework which is consists of a backbone network, a definition specified head, and the modulated rotation loss.
To enable efficiency and performance, our backbone follows RetinaNetbwhich is a representative method for one-stage object detection.
We noticed that although the eight-parameter method has been proposed for a long time, few rotated detection methods use this for parameterization.
RSDet provides two kinds of heads, a five-parameter regression one and an eight-parameter regression one, and we employ the modulated rotation loss to improve them.

\begin{figure}[tb!]
    \begin{center}
        \includegraphics[width=8cm, height=5cm]{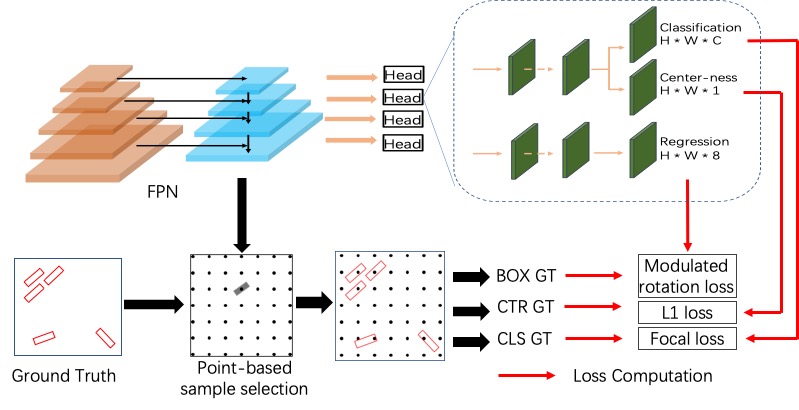}
    \end{center}
    \caption{The overview of RSDet++ framework. RSDet++ employs a point-based regression method to avoid the ignorance of tiny sample.}
    \vspace{-5pt}
    \label{fig:redet++}
    \vspace{-10pt}
\end{figure}

\subsection{Modulated Loss in  Anchor-free Method}
%子部分 问题发现
% 我们发现把我们的方法迁移到类似Dota1.5或者Dota2.发现是具有挑战性的，主要原因是这两个数据集包括了大量的长或宽小于10个像素的目标。
%所以之前的大量的anchor-based方法比如retina的效果就会大打折扣，因为基于金字塔的结构他们出于对于计算尺度和计算量的考虑，会从P3开始进行anchor的匹配。
%而在传统的FPN结构中，P3是在原始图像经过了8倍小采样的，这样对于先前提到的小目标经过小采样之后就会丧失框的结构。
%这样在anchor-based的方法在基于anchor和GT框之间的IOU来进行正负样本的挑选，会导致很多小目标的IOU低于原定阈值。
%而传统的方法为了增强对于小目标的检测能力，会使用两阶段的方法，一方面通过4倍下采样提供更多的分辨率，一方面通过第二阶段对proposal进行二次回归提升精度。
It's challenging to apply our method on benchmarks such as DOTA1.5 and DOTA2.0 since many small objects smaller than 10 pixels are difficult to be detected, and we termed this phenomenon as tiny trouble.
The tiny trouble also exists in anchor-based methods that tend to employ a pyramid architecture for performance improvement.
Moreover, they usually set up anchors from the P3 stage because the large resolution of the P1 and P2 ones will result in unaffordable computational costs.
We subsample the original image $8$ times to get the P3 stage feature,
and the objects smaller than 10 pixels will disappear after the subsampling operation.
The anchor-based methods choose the positive/negative samples by calculating the IOU between the anchor set and the ground truth set, 
and too small objects will lead to a lower IOU value which is smaller than the threshold value.
Some previous methods such as RoI transformer ~\cite{R29_ding2018learning} propose to use a two-stage detection backbone and a larger feature map such as P2 to improve the small object detection performance, 
which is computational costing and time costing.

\begin{figure}[!tb]
\centering
\subfigure[]{\includegraphics[width=2.7cm]{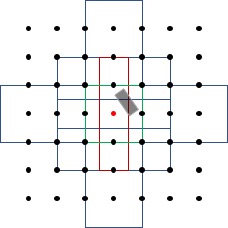}
\label{fig:anchor_compare1}} 
\subfigure[]{\includegraphics[width=2.7cm]{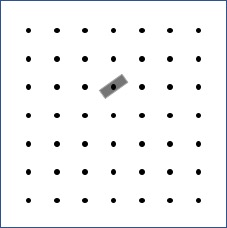}
\label{fig:anchor_compare2}}
\subfigure[]{\includegraphics[width=2.7cm]{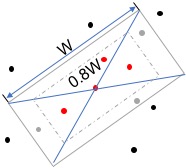}
\label{fig:anchor_compare3}}
\vspace{-5pt}
\caption{Comparison of the positive/negative sample selection process between anchor-based and point-based methods.
a) the positive/negative sample selection process in anchor-based methods by IoU computation.
b/c) the positive/negative sample selection process in point-based methods.
} %图片标题
\label{fig:anchor_compare}
\vspace{-10pt}
\end{figure}

%提出方法
%受上述任务以及anchor-free系列方法的启发，我们提出了anchor-free的基于点回归来选择正负样本的RSDet++。
%RSDet++不设定anchor，而是通过将feature map上的中心点来回归网络的八个参数，通过目标点在框中的位置来判断正负样本。
%由于点的位置不会随着下采样而丢失，所以RSDet++提高了对于小目标的检测能力，如图所示展示了anchor-based和point-based的旋转目标检测器的区别。
%值得一提的是，由于我们将anchor-based的方法修改为anchor-free，RSDet++因此获得了更加快的训练和检索速度。
Inspired by the tiny trouble and the anchor-free based method, 
we propose an RSDet++ that regresses eight-parameter of four vertexes from a point and employs the modulated rotation loss for solving ASE.
Instead of predefined anchors, RSDet++ regresses the eight-parameter values from each pixel on the feature maps and termed them as center points.
RSDet++ chooses the positive and negative samples by the relationship between center points and ground truth boxes as show in ~\ref{fig:anchor_compare3}.
As the center points won't disappear with the subsampling of the feature map, so RSDet++ can alleviate the tiny trouble better.
Specifically, the differences between the positive/negative chosen methods in anchor-based and anchor-free methods are shown in Fig.~\ref{fig:anchor_compare}.
Finally, our RSDet++ shortens the training time and provides a faster inference time.

%Although anchor-free method has been utilized in horizontal object detection method such as YOLO \cite{redmon2016you}, and FCOS \cite{R14_Tian2019FCOS},
%it has never been employed for rotated object detection.
RSDet++ employs a point-based regression method to generate the proposals,
which regresses all four vertexes coordinates from the same point.
So each point on the feature map can be viewed as an anchor point, and we need to judge whether the point belongs to a target object or not.
However, the IOU-based operation is employed to differentiate the positive sample and negative anchor in anchor-based methods, but it can't be applied to RSDet++ since no anchor exists.
So a point-based positive and negative sample selection approach is selected, given an anchor point $(x_a, y_a)$ and a target object $(x_1, y_1; x_2, y_2; x_3, y_3; x_4, y_4)$.
If the point is outside the object region, it will be termed a negative point;
and if it is inside the object region, it will be termed a positive point or an ignored one corresponding to the distance from the central point.
    \begin{equation}
     \begin{aligned}
	x_c=(x_1+x_2+x_3+x_4)/4; 
	\\
	y_c=(y_1+y_2+y_3+y_4)/4; 
	\\
	w=\sqrt{(x_1-x_2)^2+(y_1-y_2)^2}
	\\
	h=\sqrt{(x_3-x_3)^2+(y_1-y_2)^2}
	 \end{aligned}
    \end{equation}
    
%	\\
%		state=\left\{\begin{matrix}
%		&positive \quad if \;|x_a-x_c|<\alpha \times w \; and \; |y_a-y_c|<\alpha \times h
%			\\ 
%		&ignored \quad others
%			\end{matrix}\right.
As shown in the above equation, we term the anchor point $(x_a, y_a)$ as positive anchor if it satisfies $|x_a-x_c|<\alpha \times w $ and $ |y_a-y_c|<\alpha \times h$;
otherwise the point will be termed as ignored point. The above mentioned $\alpha$ is an experiential hyper-parameter, we set it as 0.8 usually. 
The above process is presented in Fig.~\ref{fig:anchor_compare}.

%方法部分
%方法的框架如下图所示：
%-------------------------------------------------------------------------
%-------------------------------------------------------------------------
%-------------------------------------------------------------------------
%-------------------------------------------------------------------------
We show the overall framework RSDet++ in Fig.~\ref{fig:redet++}, which can be summarized into three parts.
Firstly, we employ a Res50-based feature pyramid network as a baseline which has been widely employed in previous object detection methods such as RetinaNet~\cite{R25_Lin2017Focal}, R2CNN~\cite{R22_Jiang2017R2CNN}, and FCOS~\cite{R14_Tian2019FCOS}  
for its ability to focus on multi-scale objects.
Secondly, an eight-param modulated rotation loss is employed to associate with RSDet++, which can alleviate the rotation sensitivity error.
Finally, a point-based sample selection strategy is employed to avoid the tiny trouble in anchor-based methods.
%%
%%插图
%%

\section{Experiments}
To validate the effectiveness of the proposed approaches, 
we conduct extensive experiments and compare with the recent state-of-the-art methods on several challenging public benchmarks 
such as aerial images (DOTA, HRSC2016, and UCAS-AOD), scene text images (ICDAR2015).
Tensorflow is employed to implement our proposed methods on a server with V100 and 16 G memory.
The experiments in this article are initialized by ResNet50 by default unless otherwise specified.
Recall that the main contribution of this paper is to identify the problem of RSE, the introduction of modulated rotation loss, and point-based RSDet++ for tiny objects.
%Experiments are implemented by Tensorflow \cite{abadi2016tensorflow} on a server with Ubuntu 16.04, NVIDIA GTX 2080 Ti, and 11G Memory. 
%Aerial images (DOTA, HRSC2016, and UCAS-AOD), scene text images (ICDAR2015) are used for evaluation.

\subsection{Datasets and Implementation Details}

\textbf{DOTA} \cite{R18_xia2018dota}: We carry out the main experiments on the DOTA benchmark that has a total of 2,806 aerial images and 15 categories. 
The size of images in DOTA ranges from $800\times800$ pixels to $4,000\times4,000$ pixels. 
The proportions of the training set, the validation set, and the test set are respectively 1/2, 1/6, and 1/3. 
There are 188,282 instances for training and validation, and they are labeled clockwise with a quadrilateral. 
Due to the large size of a single aerial image, we divide the image into $600\times600$ pixel sub-images with a 150-pixel overlap between two neighboring ones, 
and we scale these sub-images to $800\times800$ eventually. 
There exists three different version of DOTA benchmark: DOTA-v1.0, DOTA-v1.5, and DOTA-v2.0.
DOTA-v1.5 and DOTA-v2.0 use the same images as DOTA-v1.0, but the tiny instances (smaller than 10 pixels) are included.

\textbf{ICDAR2015} \cite{karatzas2015icdar}: 
ICDAR15 is a scene text dataset that includes 1,500 images, 1000 for training and the remaining for testing. 
The size of the images in this dataset is $720\times1280$, and the source of the images is street view. 
The annotation of the text in an image is four clockwise point coordinates of a quadrangle.

\textbf{HRSC2016} \cite{liu2017a}: 
HRSC2016 is a dataset for ship detection which has a large aspect ratio and arbitrary orientation.
This dataset contains two scenarios: ship on sea and ship close inshore. 
The size of each image ranges from $300\times300$ to $1,500\times900$. 
This dataset has 1,061 images including 436 images for training, 181 for validation, and 444 for testing.

% \textbf{FDDB} \cite{jain2010fddb}: FDDB is a dataset designed for unconstrained face detection, in which faces have a wide variability of face scales, poses, and appearance. This dataset contains annotations for 5,171 faces in a set of 2,845 images taken from the faces in the Wild dataset~\cite{berg2005s}.

\textbf{UCAS-AOD} \cite{zhu2015orientation}: 
UCAS-AOD is a remote sensing dataset that contains two categories: car and plane. 
UCAS-AOD contains 1510 aerial images, each of which has approximately $659\times1,280$ pixels. 
In line with \cite{R29_ding2018learning} and \cite{R27_azimi2018towards}, we select 1110 images for training and 400 for testing randomly.

%%改到这里了，8/13/2021 5:28
\textbf{Baselines and Training Details.}
To make the experimental results more reliable, the baseline we chose is a multi-class rotated object detector based on RetinaNet, which has been verified in work \cite{R20_Yang2019R3Det}. 
During training, we use RetinaNet-Res50, RetinaNet-Res101, and RetinaNet-Res152~\cite{R25_Lin2017Focal} for experiments. 
Our network is initialized with the pre-trained ResNet50~\cite{he2016deep} for object classification in ImageNet \cite{deng2009imagenet} which is officially published by TensorFlow. 
Besides, weight decay and momentum are correspondingly 1e-4 and 0.9. 
The training epoch is 30 in total, and the number of iterations per epoch depends on the number of samples in the dataset. 
The initial learning rate is 5e-4, and the learning rate changes from 5e-5 at epoch 18 to 5e-6 at epoch 24. 
We adopt the warm-up strategy to find a suitable learning rate in the first quarter of the training epochs. 
In inference, rotating non-maximum suppression (R-NMS) is used for post-processing the final detection results.

\begin{table}[tb!]
    \centering
    \caption{Ablation experiments of $\ell_{mr}$ and predefined eight-parameter regression on DOTA benchmark.}
    \resizebox{0.4\textwidth}{!}{
        \begin{tabular}{llcc}
            \toprule
            Backbone & Loss & Regression & mAP\\
            \midrule
             resnet-50 & smooth-$\ell_1$ & five-param. & 62.14\\
             resnet-50 & $\ell_{mr}$ & five-param. & 64.49\\
             resnet-50 & smooth-$\ell_1$ & eight-param. & 65.59\\
             resnet-50 & $\ell_{mr}$ & eight-param. & \textbf{66.77}\\
            \bottomrule
        \end{tabular}}      
\label{tab:ablation_dota}
 \vspace{-10pt}
\end{table}

\subsection{Ablation Study}
\textbf{Ablation Study of Modulated Rotation Loss in 5-parameter and 8-parameter settings.}
We use the ResNet50-based RetinaNet-H as our baseline to verify the effectiveness of modulated rotation loss $\ell_{mr}$. We get a gain of 2.35\% mAP, 
when the loss function is changed from the smooth-$\ell_1$ loss to $\ell_{mr}$, as shown in Table.~\ref{tab:ablation_dota}. 
Fig.~\ref{fig:R_Qloss_vis} compares results before and after solving the RSE problem: some objects in the images are all in the boundary case where the loss function is not continuous. 
A lot of inaccurate results (see red squares in Fig.~\ref{fig:R_Qloss_vis1} and Fig.~\ref{fig:R_Qloss_vis3}) are predicted in the baseline method, 
but these do not occur after using $\ell_{mr}$ (see the same location in Fig.~\ref{fig:R_Qloss_vis2} and Fig.~\ref{fig:R_Qloss_vis4}). 
Similarly, an improvement of 1.18\% mAP is obtained after using the eight-parameter $\ell_{mr}$. 
This set of ablation experiments prove that $\ell_{mr}$ is effective for improving the rotated object detector. 
Note the number of parameters and calculations added by these two techniques are almost negligible.

\begin{figure*}
\centering
\subfigure[RSDet]
{\includegraphics[width=4cm]{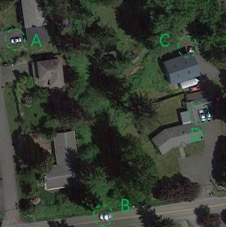}
\label{fig:small_targets1}} 
\subfigure[RSDet++]
{\includegraphics[width=4cm]{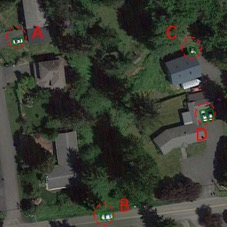}
\label{fig:small_targets2}}
\subfigure[Magnified visualization]
{\includegraphics[width=4cm]{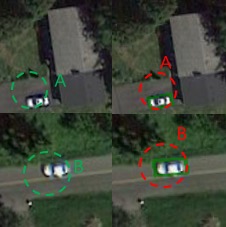}
\label{fig:small_targets3}}
 \vspace{-5pt}
\caption{
Visualization of the tiny object detection ability between our RSDet and RSDet++, 
where the circles localize the ground-truth objects, and A/B/C/D represents different objects.
For a clear comparison, we enlarge the local regions A and B in subfigure c,
and conclude that the same object neglected in RSDet can be detected in RSDet++.
} %图片标题
\label{fig:small_targets}
 \vspace{-10pt}
\end{figure*}

\textbf{Ablation Study of Modulated Rotation loss on RSDet++.}
The experiments in Table.~\ref{tab:ablation_dota} show that our modulated rotation loss can enhance the performance of existing rotated object detection methods suffered from the discontinuity of loss.
For solving the tiny trouble in anchor-based object detection methods, we introduce the point-based RSDet++ to regress the rotated bounding box from pixels on the feature map.
However, the discontinuity of loss also exists in RSDet++.
We compare the performance of RSdet++ with modulated rotation loss (RSDet++ w.mr) and RSDet++ without modulated rotation loss (RSDet++ wo.mr) in Table.~\ref{tab:mr_rsdet_plus},
the results shows that modulated rotation loss also works on RSDet++ which improve the baseline from 67.91\% to 68.24\%.

\begin{table}[tb!]
    \centering
    \caption{Ablation experiments of $\ell_{mr}$ in RSDet++ on DOTA benchmark.}
    \resizebox{0.45\textwidth}{!}{
        \begin{tabular}{ccccc}
            \toprule
            & Backbone & Loss & Regression & mAP\\
            \midrule
           RSDet wo.mr & resnet-50 & smooth-$\ell_1$ & eight-param. & 66.05\\
           RSDet w.mr & resnet-50 & $\ell_{mr}$ & eight-param. & 67.20\\
           RSDet++ wo.mr & resnet-50 & smooth-$\ell_1$ & eight-param. & 67.91\\
           RSDet++ w.mr & resnet-50 & $\ell_{mr}$ & eight-param. & 68.24\\
            \bottomrule
        \end{tabular}}      
\label{tab:mr_rsdet_plus}
 \vspace{-10pt}
\end{table}

\textbf{Ablation Study of Backbone, Data Augmentation, and Data Balance.}
Data augmentation can improve model performance effectively, 
and we use augmentations including random horizontal flipping, random vertical flipping, random image graying, and random rotation. 
Consequently, the baseline performance increased by 4.22\% to 70.79\% on DOTA. 
Moreover, data imbalance is severe in the DOTA. 
For instance, there are 76,833 ship instances in the dataset, but there are only 962 ground track fields. 
We extend samples less than 10,000 to 10,000 in each category by random copying, which brings a 0.43\% boost, and the most prominent contribution is from a small number of samples, e.g., helicopter and swimming pool. 
We also explore the impact of different backbones and conclude that a greater backbone brings more performance gains. 
Performances of the detectors based on ResNet50, ResNet101, and ResNet152 are respectively 71.22\%, 72.16\% and 73.51\%. 
Refer to Table.~\ref{tab:data_backbone} for details.

\begin{table}[tb!]
    \centering
     \caption{Ablation experiments of backbone, data augmentation and balance on DOTA. RSDet is the base model.}
    \resizebox{0.3\textwidth}{!}{
    \begin{tabular}{lccc}
        \toprule
         Backbone&Data Aug&Balance&mAP\\
        \midrule
        resnet-50 &  &  & 66.77\\
         resnet-50 & $\checkmark$ &  & 70.79\\
         resnet-50 & $\checkmark$ & $\checkmark$ & 71.22\\
         resnet-101 & $\checkmark$ & $\checkmark$ & 72.16\\
         resnet-152 & $\checkmark$ & $\checkmark$ & \textbf{73.51}\\
        \bottomrule
    \end{tabular}}
    \label{tab:data_backbone}
     \vspace{-10pt}
\end{table}

\textbf{Ablation Study of Two-stage Detectors as Base Model.}
To prove that $\ell_{mr}$ can cooperate with other existing frameworks, we apply $\ell_{mr}$ to the  two-stage methods such as RSDet-II and Faster RCNN. 
By comparison, the $\ell_{mr}$ achieves consistent improvement on RSDet-II and Faster RCNN. 
We conduct Faster RCNN based on five-parameter and eight-parameter systems, while RSDet-II only conducts on a five-parameter system. 
Taking Faster RCNN as an example, it outperforms baseline by 1.6\% and 2.84\% in mAP after cooperating with $\ell_{mr}$ in five-parameter and eight-parameter, respectively. 
The results of RSDet-II are presented in Table.~\ref{tab:dota_sota}, which reach state-of-the-art on the DOTA benchmark.

\textbf{Ablation Study of RSDet and RSDet++.}
We compare RSDet and RSDet++ as shown in Table.~\ref{tab:mr_rsdet_plus},
and select the same backbone ResNet50 and eight-parameter regression system for a fair comparison.
We note that the performance of RSDet++ both with/without modulated rotation loss are both significantly better than that of RSDet.
When comparing the methods without modulated rotation loss, RSDet++ receives 67.91\% mAP that is 1.86\% higher than RSDet; 
when comparing the methods with modulated rotation loss, RSDet++ receives 68.24\% mAP that is 1.04\% higher than RSDet.
The above experiments show that when we improve the performance of our methods on tiny objects, point-based methods can achieve better performance than anchor-based methods with a more efficient inference process.
We explain the gap  that the RSDet++ uses a point-based sample selection mode which is more suitable for tiny objects than the traditional IOU-based sample selection mode in anchor-based methods.
Moreover, anchor-free methods can provide a more efficient inference process.

In Fig.~\ref{fig:small_targets}, we also visualize the detection results using the anchor-based RSDet and point-free RSDet++ respectively. 
We can find that RSDet++ can predict the bounding box of small objects while RSDet can't.

\begin{figure}[!tb]
\centering
\subfigure[Loss curves (five-param.)]{\includegraphics[width=4.0cm, height=3cm]{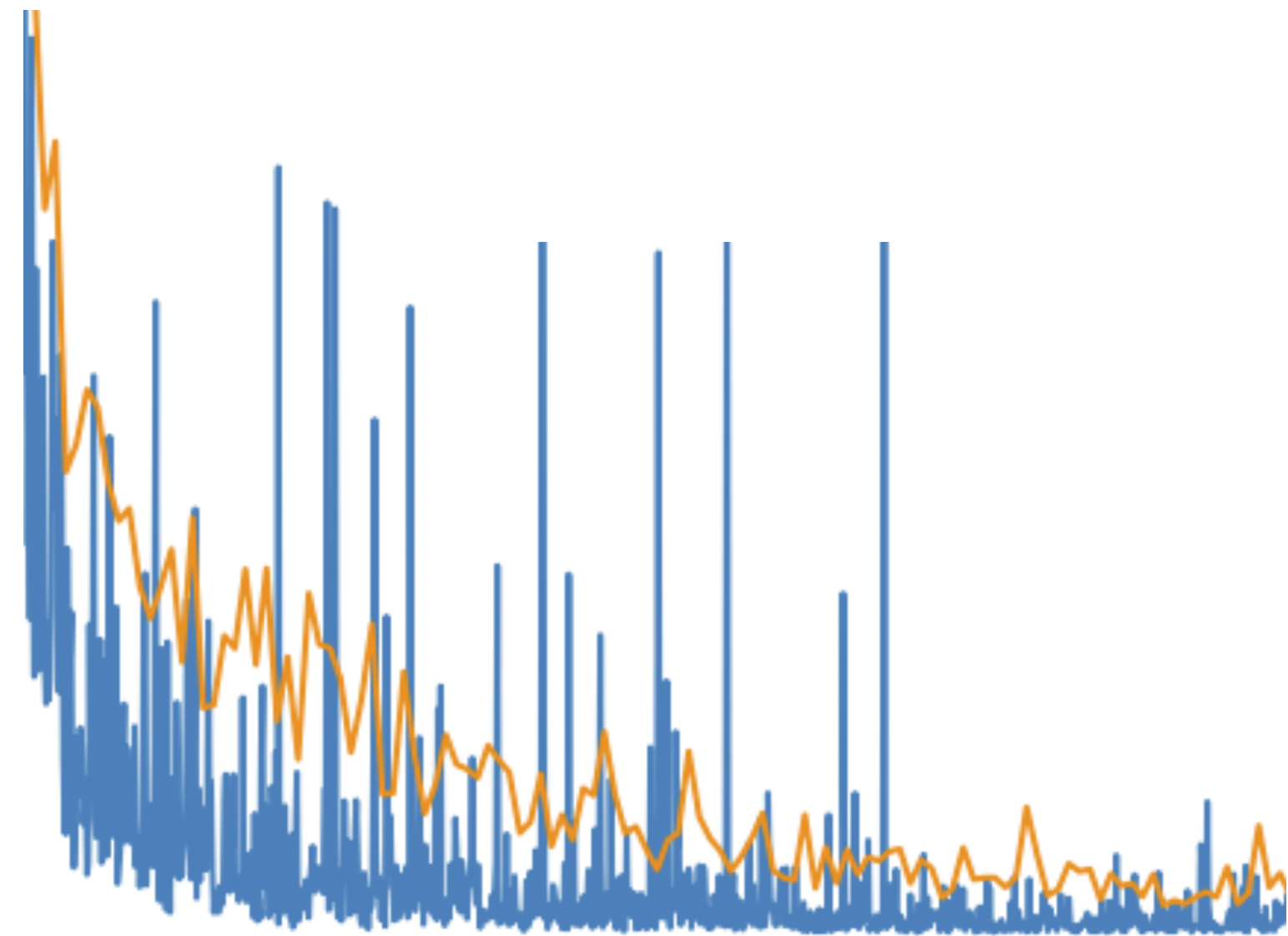}} 
\label{fig:fig6_a}
\subfigure[Loss curves (eight-param.)]{\includegraphics[width=4.0cm,height=3cm]{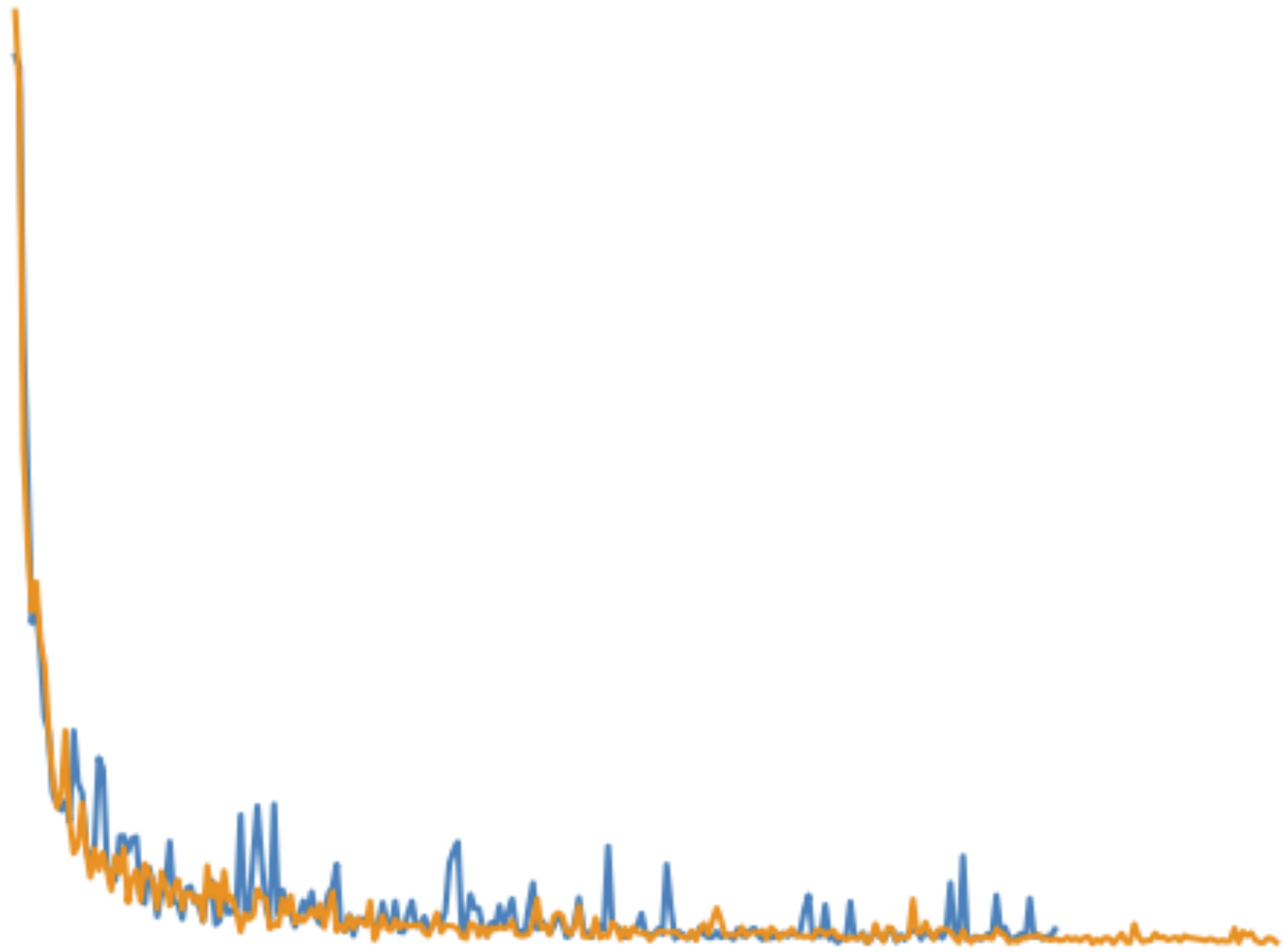}}
\label{fig:fig6_a}
 \vspace{-5pt}
\caption{Comparisons of loss curves during training: where blue for L1-loss and yellow for our loss.} %图片标题
\label{fig:loss_curve}
 \vspace{-10pt}
\end{figure}

\subsection{Further Analysis}

\begin{figure*}[!tb]
\centering
\subfigure[Vehicles]{\includegraphics[width=4cm,height=4cm]{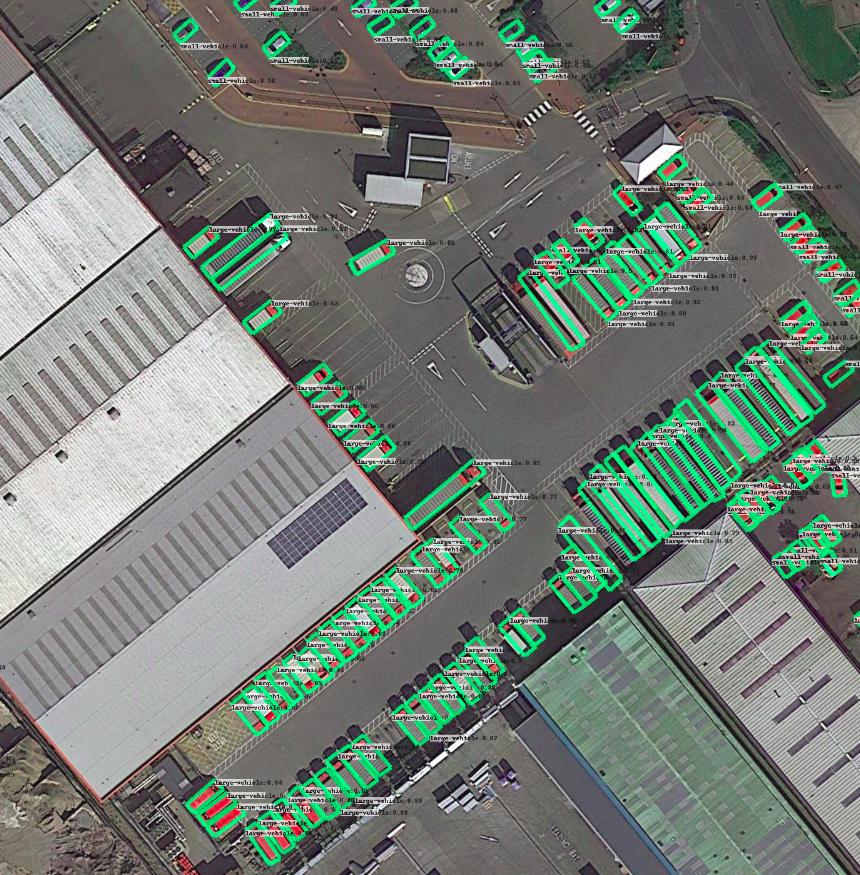}}
\subfigure[Swimming pool]{\includegraphics[width=4cm,height=4cm]{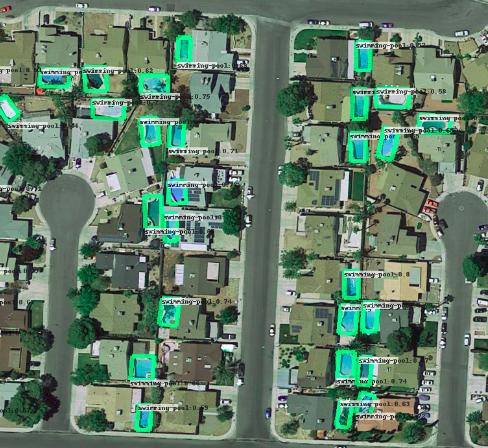}}
\subfigure[Tennis field]{\includegraphics[width=4cm,height=4cm]{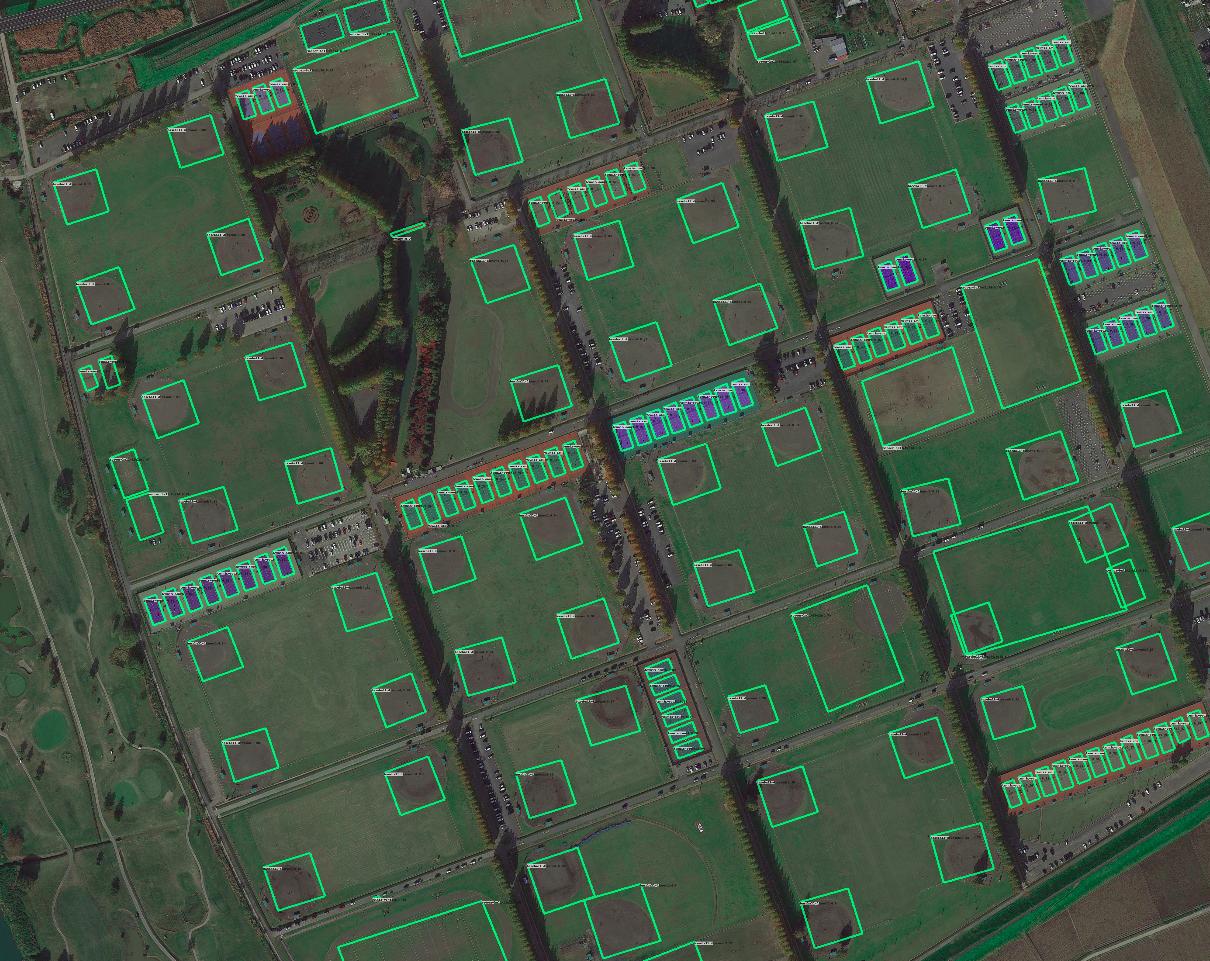}}
\subfigure[Text in signs]{\includegraphics[width=4cm,height=4cm]{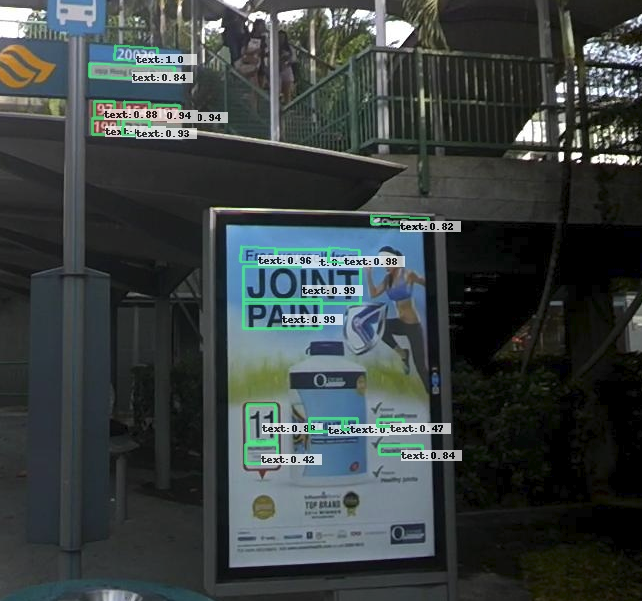}}
\\
\subfigure[Harbor]{\includegraphics[width=4cm,height=4cm]{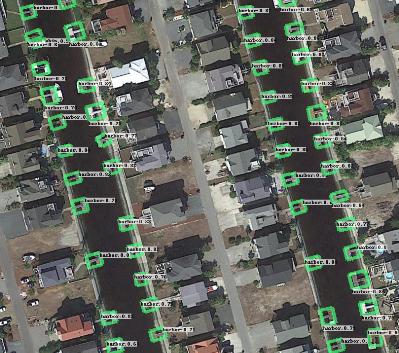}}
\subfigure[Storage tank]{\includegraphics[width=4cm,height=4cm]{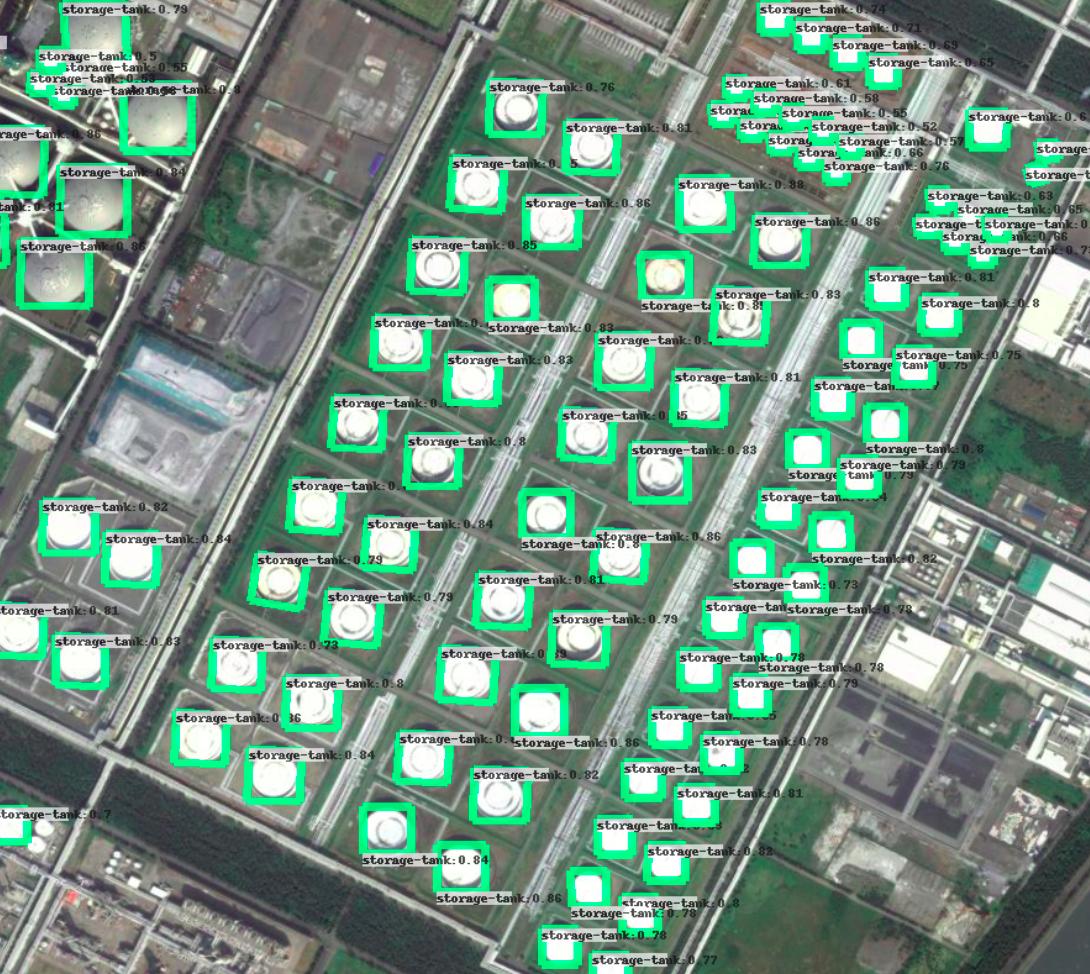}}
\subfigure[Harbor and ship]{\includegraphics[width=4cm,height=4cm]{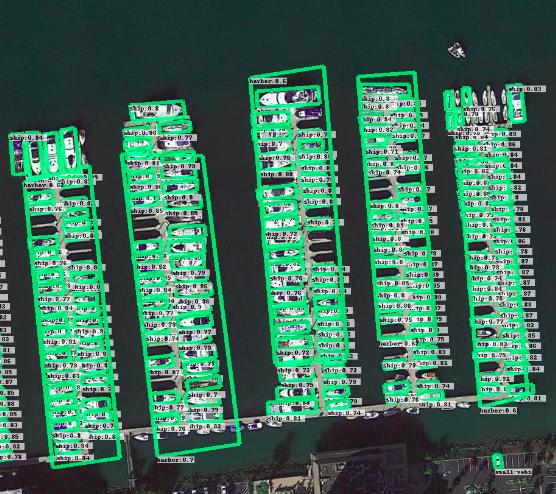}}
\subfigure[Text in elevator]{\includegraphics[width=4cm,height=4cm]{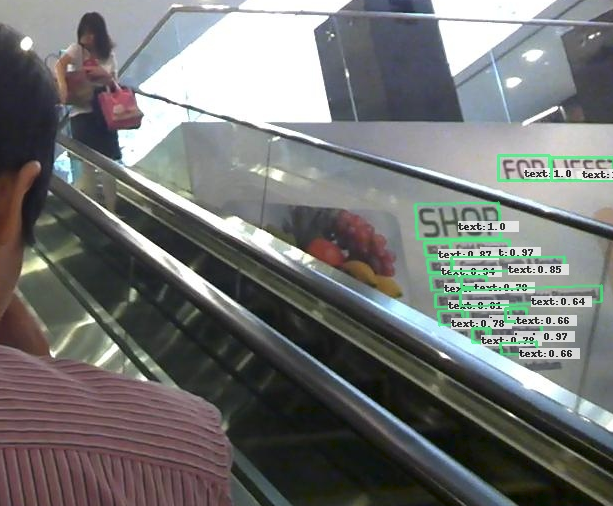}}
\caption{Detection results on DOTA and ICDAR15.} %图片标题
 \vspace{-5pt}
\label{fig:dota_vis}
 \vspace{-10pt}
\end{figure*}

\textbf{Training Stability.}
We have analyzed that the discontinuity of loss affects the training stability and the detection performance. 
Although the detection performances of our methods are verified through mAP, we have not proven the stability improvement of model training brought by our techniques. 
To this end, we plot the training loss curves using models including RetinaNet-H ($\ell^{5p}$), RSDet-I ($\ell^{5p}_{mr}$), RetinaNet-H ($\ell^{8p}$), and RSDet-I ($\ell^{8p}_{mr}$), as shown in Fig.~\ref{fig:loss_curve}. 
The training process converges more stable when associating with a modulated rotation loss.

\textbf{Comparison with Similar Methods.}
Although we introduce the concept of RSE for the first time, it is worth noting that some previous articles have also mentioned similar problems. 
In \cite{R18_xia2018dota}, they use a 180-degree definition to eliminate the loss burst caused by the exchangeability of height and width. 
While related works \cite{R30_ma2018arbitrary, bao2019single} use periodic trigonometric functions to eliminate the effects of the angular periodicity. 
SCRDet \cite{R28_Yang2018SCRDet} proposes IoU-smooth-$\ell_1$ loss to solve the boundary discontinuity. However, these methods are limited and do not completely solve the RSE problem. 
Our method yields the most promising results when comparing with these methods as shown in Table.~\ref{tab:similar_methods}.

\begin{table}[tb!]
    \centering
     \caption{Ablation study using the proposed techniques on DOTA. RetinaNet-H\cite{R20_Yang2019R3Det} is the base model.}
    \resizebox{0.38\textwidth}{!}{
    \begin{tabular}{llc}
        \toprule
        Loss & Regression & mAP\\
        \midrule
        smooth-$\ell_1$ & five-param. $[-\frac{\pi}{2},0)$ & 62.14\\
        smooth-$\ell_1$\cite{R18_xia2018dota} & five-param. $[-\pi,0)$  & 62.39\\
        IoU-smooth-$\ell_1$ \cite{R28_Yang2018SCRDet} & five-param. $[-\frac{\pi}{2},0)$ & 62.69\\
        $\ell_{mr}$ & five-param. $[-\frac{\pi}{2},0)$ & 64.49 \\
        smooth-$\ell_1$ & eight-param. & 65.59\\
        $\ell_{mr}$ & eight-param. & \textbf{66.77}\\
        \bottomrule
    \end{tabular}}  
\label{tab:similar_methods}
 \vspace{-10pt}
\end{table}

\begin{table}[tb!]
    \centering
     \caption{Performances of $\ell_{mr}$ and eight-para. regression on ICDAR2015 and HRSC2016. Use RetinaNet-H \cite{R20_Yang2019R3Det} as base model, and ResNet152 backbone.}
    \resizebox{0.38\textwidth}{!}{
    \begin{tabular}{llcc}
        \toprule
         Loss & Regression& ICDAR2015 & HRSC2016\\
        \midrule
         smooth-$\ell_1$ & five-param. & 76.8 & 82.4\\
         $\ell_{mr}$ & five-param. & 79.6 & 83.6\\
          smooth-$\ell_1$ & eight-param. & 81.2 & 85.4\\
         $\ell_{mr}$ & eight-param. & \textbf{83.2} & \textbf{86.5}\\
         \bottomrule
    \end{tabular}}
    \label{tab:hrsc2016_icdar2015}
     \vspace{-10pt}
\end{table}

\begin{table}[tb!]
    \centering
    \caption{Performance on DOTA1.5 and DOTA2.0.}
    \resizebox{0.38\textwidth}{!}{
    \begin{tabular}{cccc}
        \toprule
        Method &   Box Def. & DOTA-V1.5& DOTA-v2.0 \\
        \midrule
        IOU-smooth L1 \cite{R28_Yang2018SCRDet} & $D_{oc}$ & 59.16&46.31 \\
        RIDet \cite{ming2021optimization}  & quad. & 58.91&45.35 \\
        CSL \cite{yang2020arbitrary} & $D_{oc}$ & 58.55&43.34 \\
        DCL \cite{DBLP:journals/corr/abs-2011-09670} & $D_{oc}$ & 59.38&45.46 \\
        GWD \cite{yang2021rethinking} & $D_{oc}$ & 60.03&46.65 \\
        KLD  \cite{yang2021learning} & $D_{oc}$ & 62.50&47.69 \\
        \hline
        RSDet  & quad. & 61.42&46.71 \\
        RSDet++  & quad.& 62.18&48.81 \\
        \bottomrule
\end{tabular}}    
\label{tab:dota12}
\end{table}

\begin{table}[tb!]
    \caption{
    Accuracy comparison between different rotation detectors on DOTA dataset. 
    The bold \textbf{\color{red}{red}} fonts indicate the top one performances respectively. 
    $D_{oc}$ and $D_{le}$ represent OpenCV Definition $\left(\theta \in [-90^\circ, 0^\circ)\right)$ and Long Edge Definition $\left(\theta \in [-90^\circ, 90^\circ)\right)$ of RBox.}
    \label{tab:high_acc}
    \centering
    \resizebox{0.38\textwidth}{!}{
    \begin{tabular}{c|c|ccc}
        \toprule
        Method & Box Def. & \multicolumn{3}{c|}{v1.0 train/val}  \\
         & &  AP$_{50}$ & AP$_{75}$ & AP$_{50:95}$\\
        % & & & BR$^\dagger$ (2.93) & SV$^\dagger$ (1.72) & LV$^\dagger$ (3.45) & SH$^\dagger$ (2.4) & HA$^\dagger$ (3.92) & ST$^\ddagger$ (1.) & RA$^\ddagger$ (1.) & 7-mAP$_{50}$ & mAP$_{50}$ & mAP$_{50}$ & mAP$_{75}$ & mAP$_{50:95}$\\
        \hline
         Retina Baseline & $D_{oc}$  & 64.70 & 32.31 & 34.50 \\
         Retina Baseline & $D_{le}$  & 62.21 & 26.06 & 31.49  \\
         IoU-Smooth L1 \cite{R28_Yang2018SCRDet} & $D_{oc}$ & 64.61 & 34.17 & 36.23\\
         CSL \cite{yang2020arbitrary} & $D_{le}$ & 64.40 & 32.58 & 35.04 \\
         DCL (BCL) \cite{DBLP:journals/corr/abs-2011-09670} & $D_{le}$  & 65.93 & 35.66 & 36.71  \\
         \hline
          Modulated Loss (ours) & $D_{oc}$  & 63.50 & 33.32 & 34.61 \\
         Modulated Loss (ours)& Quad.  & 65.15 & \textbf{\color{red}{40.59}} & \textbf{\color{red}{39.12}} \\
        \bottomrule
    \end{tabular}}
\end{table}

\begin{table*}[tb!]
    \centering
        \caption{Detection accuracy on different objects and overall performances with the state-of-the-art methods on DOTA. The short names for categories are defined as (abbreviation-full name): PL-Plane, BD-Baseball diamond, BR-Bridge, GTF-Ground field track, SV-Small vehicle, LV-Large vehicle, SH-Ship, TC-Tennis court, BC-Basketball court, ST-Storage tank, SBF-Soccer-ball field, RA-Roundabout, HA-Harbor, SP-Swimming pool, and HC-Helicopter. BB means Backbone.}
    \resizebox{1.\textwidth}{!}{
    \begin{tabular}{lcccccccccccccccccc}
        \toprule
        Method&BB&Reg.&PL&BD&BR&GTF&SV&LV&SH&TC&BC&ST&SBF&RA&HA&SP&HC&mAP\\
		\midrule
		\textbf{Two-stage} \\
        \midrule
        % FR-O~\cite{R18_xia2018dota} &ResNet101 & 5-para & & 79.09 & 69.12 & 17.17 & 63.49 & 34.20 & 37.16 & 36.20 & 89.19 & 69.60 & 58.96 & 49.4 & 52.52 & 46.69 & 44.80 & 46.30 & 52.93\\
        IENet \cite{lin2019ienet}  &R-101 & 6p & 80.2 & 64.5 & 39.8 & 32.0 & 49.7 & 65.0 & 52.6 & 81.5 & 44.7 & 78.5 & 46.5 & 56.7 & 64.4 & 64.2 & 36.8 & 57.1 \\
        R-DFPN \cite{R21_yang2018automatic} &R-101 & 5p  & 80.9 & 65.8 & 33.8 & 58.9 & 55.8 & 50.9 & 54.8 & 90.3 & 66.3 & 68.7 & 48.7 & 51.8 & 55.1 & 51.3 & 35.9 & 57.9 \\
        R$^2$CNN \cite{R22_Jiang2017R2CNN} &R-101 & 5p & 80.9 & 65.7 & 35.3 & 67.4 & 59.9 & 50.9 & 55.8 & 90.7 & 66.9 & 72.4 & 55.1 & 52.2 & 55.1 & 53.4 & 48.2 & 60.7\\
        RRPN \cite{R30_ma2018arbitrary} &R-101 & 5p & 88.5 & 71.2 & 31.7 & 59.3 & 51.9 & 56.2 & 57.3 & 90.8 & 72.8 & 67.4 & 56.7 & 52.8 & 53.1 & 51.9 & 53.6 & 61.0\\
        
        % RetinaNet-H+ResNet50 \cite{R20_Yang2019R3Det} & 88.9 & 74.5 & 40.1 & 58.0 & 63.1 & 50.6 & 63.6 & 90.9 & 77.9 & 76.4 & 48.3 & 55.9 & 50.7 & 60.2 & 34.2 & 62.2 \\
        
        % RetinaNet-R+ResNet50 \cite{R20_Yang2019R3Det} & 88.9 & 67.7 & 33.6 & 56.8 & 66.1 & 73.3 & 75.2 & 90.9 & 74.0 & 75.1 & 43.8 & 56.7 & 51.1 & 55.7 & 21.5 & 62.0 \\
        
        ICN \cite{R27_azimi2018towards} &R-101 & 5p & 81.4 & 74.3 & 47.7 & 70.3 & 64.9 & 67.8 & 70.0 & 90.8 & 79.1 & 78.2 & 53.6 & 62.9 & 67.0 & 64.2 & 50.2 & 68.2\\
        RoI Trans. \cite{R29_ding2018learning} &R-101 & 5p & 88.6 & 78.5 & 43.4 & 75.9 & 68.8 & 73.7 & 83.6 & 90.7 & 77.3 & 81.5 & 58.4 & 53.5 & 62.8 & 58.9 & 47.7 & 69.6\\
        % P-RSDet \cite{zhou2020objects} & 89.02 & 73.65 & 47.33 & 72.03 & 70.58 & 73.71 & 72.76 & 90.82 & 80.12 & 81.32 & 59.45 & 57.87 & 60.79 & 65.21 & 52.59 & 69.82 \\  
        % O$^2$-DNet \cite{wei2019oriented} & 89.31 & 82.14 & 47.33 & 61.21 & \textbf{71.32} & 74.03 & 78.62 & 90.76 & 82.23 & 81.36 & 60.93 & 60.17 & 58.21 & 66.98 & 61.03 & 71.04 \\
        CAD-Net \cite{zhang2019cad} &R-101 & 5p & 87.8 & 82.4 & 49.4 & 73.5 & 71.1 & 63.5 & 76.7 & 90.9 & 79.2 & 73.3 & 48.4 & 60.9 & 62.0 & 67.0 & 62.2 & 69.9 \\
        SCRDet \cite{R28_Yang2018SCRDet}&R-101 & 5p &  89.9 & 80.7 & 52.1 & 68.4 & 68.4 & 60.3 & 72.4 & 90.9 & 87.9 & 86.9 & 65.0 & 66.7 & 66.3 & 68.2 & 65.2 & 72.6 \\
        Gliding Ver. \cite{2020Gliding} &R-101 & 9p & 89.6 & 85.0 & 52.3 & \textbf{77.3} & 73.0 & 73.1 & 86.8 & 90.7 & 79.0 & 86.8 & 59.6 & 70.9 & 72.9 & 70.9 & 57.3 & 75.0 \\
		%Mask OBB &R-101 & pixel & 89.6 & \textbf{85.9} & 54.2 & 72.9 & 76.5 & 74.2 & 85.6 & 89.9 & 83.8 & 86.5 & 54.9 & \textbf{69.6} & 73.9 & 69.1 & 63.3 & 75.3 \\
		FFA \cite{fu2020rotation} &R-101 & 5p & 90.1 & 82.7 & 54.2 & 75.2 & 71.0 & \textbf{79.9} & 83.5 & 90.7 & 83.9 & 84.6 & 61.2 & 68.0 & 70.7 & 76.0 & 63.7 & 75.7 \\
		APE \cite{zhu2020adaptive} &R-101 & 5p & 89.9 & 83.6 & 53.4 & 76.0 & 74.0 & 77.2 & 79.5 & 90.8 & 87.2 & 84.5 & 67.7 & 60.3 & 74.6 & 71.8 & 65.6 & 75.8 \\
		CenterMap \cite{wang2020learning} &R-101 & 5p & 89.8 & 84.4 & 54.6 & 70.3 & \textbf{77.7} & 78.3 & \textbf{87.2} & 90.7 & 84.9 & 85.3 & 56.5 & 69.2 & 74.1 & 71.6 & 66.1 & 76.0\\
		RSDet-II  &R-152 & 8p & 89.9& \textbf{84.5} & 53.8 & 74.4 & 71.5 & 78.3 & 78.1 & 91.1 & 87.4 & \textbf{86.9} & 65.6 & 65.2 & \textbf{75.4} & \textbf{79.7} & 63.3 & \textbf{76.3} \\
		\hline
		\textbf{One-stage} & \multicolumn{16}{c}{} \\
		\hline	
        P-RSDet \cite{zhou2020objects} &R-101 & 3p & 89.0 & 73.7 & 47.3 & 72.0 & 70.6 & 73.7 & 72.8 & 90.8 & 80.1 & 81.3 & 59.5 & 57.9 & 60.8 & 65.2 & 52.6 & 69.8 \\
		O$^2$-DNet \cite{wei2019oriented} &H-104 & 10p & 89.3 & 82.1 & 47.3 & 61.2 & 71.3 & 74.0 & 78.6 & 90.8 & 82.2 & 81.4 & 60.9 & 60.2 & 58.2 & 66.9 & 61.0 & 71.0 \\
		DRN \cite{2020Dynamic} &H-104 & 5p & 89.7 & 82.3 & 47.2 & 64.1 & 76.2 & 74.4 & 85.8 & 90.6 & 86.2 & 84.9 & 57.7 & 61.9 & 69.3 & 69.6 & 58.5 & 73.2 \\
        R$^3$Det \cite{R20_Yang2019R3Det} &R-152 & 5p & 89.5 & 81.2 & 50.5 & 66.1 & 70.9 & 78.7 & 78.2 & 90.8 & 85.3 & 84.2 & 61.8 & 63.8 & 68.2 & 69.8 & 67.2 & 73.7\\		
        % RSDet+ResNet50 (ours) & 89.3 & 82.7 & 47.7 & 63.9 & 66.8 & 62.0 & 67.3 & 90.8 & 85.3 & 82.4 & 62.3 & 62.4 & 65.7 & 68.6 & 64.6 & 70.8\\
        % RSDet+ResNet101 (ours) & 89.8 & 82.9 & 48.6 & 65.2 & 69.5 & 70.1 & 70.2 & 90.5 & 85.6 & 83.4 & 62.5 & 63.9 & 65.6 & 67.2 & \textbf{68.0} & 72.2\\
        % RSDet+ResNet152 (ours) & \textbf{90.2} & \textbf{83.5} & 53.6 & 70.1 & 64.6 & 79.4 & 67.3 & 91.0 & \textbf{88.3} & 82.5 & 64.1 & \textbf{68.7} & 62.8 & 69.5 & 66.9 & 73.5\\
        % RSDet-I
        % & 90.1 & 82.0 & \textbf{53.8} & 68.5 & 70.2 & \textbf{78.7} & 73.6 & \textbf{91.2} & 87.1 & 84.7 & 64.3 & 68.2 & 66.1 & \textbf{69.3} & 63.7 & \textbf{74.1}\\
        RSDet-I &R-152 & 8p & \textbf{90.0}& 83.9 & \textbf{54.7} & 69.9 & 70.6 & 79.6 & 75.4 & \textbf{91.2} & \textbf{88.0} & 85.6 & 65.2 & \textbf{69.2} & 67.0 & 70.2 & 64.6 & 75.0 \\
        RSDet++&R-152 & 8p& 86.8&82.7&54.6& 71.7&76.0&71.2&83.5&87.4&83.4& 85.3&\textbf{72.4}&62.9&70.9&72.3&\textbf{70.4}& \textbf{75.4}\\
        \bottomrule
    \end{tabular}}
    \label{tab:dota_sota}
\end{table*}

\textbf{Performances on Other Datasets.}
We further do experiments on ICDAR2015, and HRSC2016 as shown in Table.~\ref{tab:hrsc2016_icdar2015}. For ICDAR2015, there are rich existing methods such as R$^2$CNN, Deep direct regression~\cite{he2017deep} and FOTS~\cite{liu2018fots}, and the current state-of-art has reached 91.67\%. 
Although they use many text-based tricks, we notice that they are also not aware of the rotation sensitivity error. 
Therefore, we conduct some verification experiments based on $\ell_{mr}$, and we achieve improvements on both datasets. 
Our detector achieves competitive results and shows its generalization on scene text data. 
Besides, we also experiment with the modulated loss on HRSC2016, and the experimental results are comparable to state-of-art.

\textbf{Performance of RSDet++ on DOTA1.5 and DOTA2.0.}

Table~\ref{tab:dota12} reports the performance of RSDet++ and other previous state-of-the-art rotation object detection methods on DOTA-v1.5 and DOTA-v2.0
such as IOU-smooth L1, RIDet, CSL, DCL, GWD, and KLD.
Without bells and whistles, RSDet++ can achieve competitive performance when comparing with other anchor-based methods.
For example, RSDet++ employs an anchor-free backbone which is more efficient and outperforms the second-best method KLD by 1.12\% mAP on DOTA-v2.0 while also achieves competitive results with KLD on DOTA-v1.5.

\textbf{High Precision Protocol of Modulated Loss.}
As aforementioned, our proposed RSDet++ show better performance on tiny objects than pervious anchor-based methods.
Moreover, we evaluate RSDet++ on some high precision protocols such as $AP_{75}$ and $AP_{50:95}$ in Table~\ref{tab:high_acc}, 
and find that quadrilateral-based Modulated Loss achieves higher performance than other methods such as IoU-Smooth L1 \cite{R28_Yang2018SCRDet}, CSL \cite{yang2020arbitrary}, and DCL \cite{DBLP:journals/corr/abs-2011-09670}.
For example, the quadrilateral-based Modulated Loss achieves 40.59\% on $AP_{75}$ which outperforms the DCL by a large marge 4.93\%,
a similar result can be achieved on  $AP_{50:95}$ which outperforms the DCL by a large marge 2.41\%.

\subsection{Overall Evaluation}
The results on DOTA1.0 are shown in Table~\ref{tab:dota_sota}. The compared methods include i) one-stage methods: P-RSDet \cite{zhou2020objects} , O$^2$-DNet \cite{wei2019oriented} , DRN \cite{2020Dynamic}, R$^3$Det \cite{R20_Yang2019R3Det}  ii) two-stage methods: R$^{2}$CNN \cite{R22_Jiang2017R2CNN}, Gliding Vertex \cite{2020Gliding}, FFA \cite{fu2020rotation}, APE \cite{zhu2020adaptive}, CenterMap OBB \cite{wang2020learning}. The results of DOTA are all obtained by submitting predictions to the official evaluation server. 
We apply $\ell_{mr}$ to the one-stage and two-stage approaches, and call them RSDet-I and RSDet-II, respectively. 
In the one-stage methods, RSDet-I obtained 75.03$\%$ mAP, which is 1.29\% higher than the existing best method (R$^3$Det). 
And in the two-stage methods, RSDet-II obtained 76.34$\%$ mAP, which is 0.31\% higher than CenterMap OBB, 1.28\% higher than Gliding Vertex. 
Table.~\ref{tab:ucas-aod} gives the comparison on the UCAS-AOD dataset, where our method achieves 96.50\% for the OBB task that outperforms all the published methods. 
Moreover, the number of parameters and calculations added by our techniques are almost negligible, and they can be applied to all region-based rotation detection algorithms. 
Visualization results are shown in Fig.~\ref{fig:dota_vis}.

\begin{table}[tb!]
    \centering
    \caption{Performance evaluation on UCAS-AOD dataset.}
    \resizebox{0.38\textwidth}{!}{
    \begin{tabular}{lcclc}
        \toprule
        Method &   Plane && Car\quad\quad& mAP \\
        \midrule
        YOLOv2 \cite{redmon2016you}  & 96.60 && 79.20 \quad\quad& 87.90 \\
        R-DFPN \cite{R21_yang2018automatic}  & 95.90 && 82.50 \quad\quad& 89.20 \\
        DRBox \cite{liu2017learning}  & 94.90& & 85.00 \quad\quad& 89.95 \\
        S$^2$ARN \cite{bao2019single}  & 97.60 && 92.20 \quad\quad& 94.90 \\
        RetinaNet-H \cite{R20_Yang2019R3Det} & 97.34 && 93.60  \quad\quad& 95.47 \\
        ICN \cite{R27_azimi2018towards} & - && -  \quad\quad & 95.67\\
        FADet \cite{li2019feature}  & 98.69 && 92.72 \quad\quad& 95.71 \\
        R$^3$Det \cite{R20_Yang2019R3Det}  & 98.20 && 94.14 \quad\quad& 96.17 \\
        \hline
        Ours   & \textbf{98.04} && \textbf{94.97} \quad\quad& \textbf{96.50}\\
        \bottomrule
\end{tabular}}    
\label{tab:ucas-aod}
\end{table}

\section{Conclusion}
In this paper, the issue of rotation sensitivity error (RSE) is formally identified and formulated for region-based rotated object detectors. 
RSE mainly refers to the discontinuity of loss which caused by the contradiction between the definition of the rotated bounding box and the loss function. 
A novel modulated rotation loss $\ell_{mr}$ is proposed to address the discontinuity of loss in five-parameter and eight-parameter methods. 
To prove the effectiveness of modulated loss, we conduct experiments based on one-stage and two-stage methods respectively. 
Moreover, we further propose a RSDet++ to improve the performance on tiny objects based on a point-based bounding box regression approach.
Extensive experiments demonstrate that RSDet achieves state-of-art performance on the DOTA1.0 benchmark and RSDet++ performs better on DOTA1.5 and DOTA2.0 which focus more on tiny objects.

\section{ACKNOWLEDGMENT}
This research is supported by The Commercialization of Research Findings Fund of Inner Mongolia Autonomous Region (2020CG0075).
The author Wen Qian is supported by Multidimensional Data Analysis Laboratory, Institute of Automation, Chinese Academy of Sciences.

\bibliographystyle{IEEEtran}
\bibliography{IEEEabrv,new}

\begin{thebibliography}{10}
\providecommand{\url}[1]{#1}
\csname url@samestyle\endcsname
\providecommand{\newblock}{\relax}
\providecommand{\bibinfo}[2]{#2}
\providecommand{\BIBentrySTDinterwordspacing}{\spaceskip=0pt\relax}
\providecommand{\BIBentryALTinterwordstretchfactor}{4}
\providecommand{\BIBentryALTinterwordspacing}{\spaceskip=\fontdimen2\font plus
\BIBentryALTinterwordstretchfactor\fontdimen3\font minus
  \fontdimen4\font\relax}
\providecommand{\BIBforeignlanguage}[2]{{%
\expandafter\ifx\csname l@#1\endcsname\relax
\typeout{** WARNING: IEEEtran.bst: No hyphenation pattern has been}%
\typeout{** loaded for the language `#1'. Using the pattern for}%
\typeout{** the default language instead.}%
\else
\language=\csname l@#1\endcsname
\fi
#2}}
\providecommand{\BIBdecl}{\relax}
\BIBdecl

\bibitem{R18_xia2018dota}
G.-S. Xia, X.~Bai, J.~Ding, Z.~Zhu, S.~Belongie, J.~Luo, M.~Datcu, M.~Pelillo,
  and L.~Zhang, ``Dota: A large-scale dataset for object detection in aerial
  images,'' in \emph{Proceedings of the IEEE conference on computer vision and
  pattern recognition}, 2018, pp. 3974--3983.

\bibitem{zhu2015orientation}
H.~Zhu, X.~Chen, W.~Dai, K.~Fu, Q.~Ye, and J.~Jiao, ``Orientation robust object
  detection in aerial images using deep convolutional neural network,'' in
  \emph{IEEE International Conference on Image Processing}.\hskip 1em plus
  0.5em minus 0.4em\relax IEEE, 2015, pp. 3735--3739.

\bibitem{liu2017a}
Z.~Liu, L.~Yuan, L.~Weng, and Y.~Yang, ``A high resolution optical satellite
  image dataset for ship recognition and some new baselines,'' pp. 324--331,
  2017.

\bibitem{karatzas2015icdar}
D.~Karatzas, L.~Gomez-Bigorda, A.~Nicolaou, S.~Ghosh, A.~Bagdanov, M.~Iwamura,
  J.~Matas, L.~Neumann, V.~R. Chandrasekhar, S.~Lu \emph{et~al.}, ``Icdar 2015
  competition on robust reading,'' in \emph{2015 13th International Conference
  on Document Analysis and Recognition}.\hskip 1em plus 0.5em minus 0.4em\relax
  IEEE, 2015, pp. 1156--1160.

\bibitem{gomez2017icdar2017}
R.~Gomez, B.~Shi, L.~Gomez, L.~Numann, A.~Veit, J.~Matas, S.~Belongie, and
  D.~Karatzas, ``Icdar2017 robust reading challenge on coco-text,'' in
  \emph{2017 14th IAPR International Conference on Document Analysis and
  Recognition}, vol.~1.\hskip 1em plus 0.5em minus 0.4em\relax IEEE, 2017, pp.
  1435--1443.

\bibitem{DBLP:conf/aaai/Qian0PYG21}
\BIBentryALTinterwordspacing
W.~Qian, X.~Yang, S.~Peng, J.~Yan, and Y.~Guo, ``Learning modulated loss for
  rotated object detection,'' in \emph{Thirty-Fifth {AAAI} Conference on
  Artificial Intelligence, {AAAI} 2021, Thirty-Third Conference on Innovative
  Applications of Artificial Intelligence, {IAAI} 2021, The Eleventh Symposium
  on Educational Advances in Artificial Intelligence, {EAAI} 2021, Virtual
  Event, February 2-9, 2021}.\hskip 1em plus 0.5em minus 0.4em\relax {AAAI}
  Press, 2021, pp. 2458--2466. [Online]. Available:
  \url{https://ojs.aaai.org/index.php/AAAI/article/view/16347}
\BIBentrySTDinterwordspacing

\bibitem{girshick2014rich}
R.~Girshick, J.~Donahue, T.~Darrell, and J.~Malik, ``Rich feature hierarchies
  for accurate object detection and semantic segmentation,'' in
  \emph{Proceedings of the IEEE conference on computer vision and pattern
  recognition}, 2014, pp. 580--587.

\bibitem{girshick2015fast}
R.~Girshick, ``Fast r-cnn,'' in \emph{The IEEE International Conference on
  Computer Vision}, 2015, pp. 1440--1448.

\bibitem{R16_Ren2015Faster}
S.~Ren, K.~He, R.~B. Girshick, and J.~Sun, ``Faster {R-CNN:} towards real-time
  object detection with region proposal networks,'' \emph{{IEEE} Trans. Pattern
  Anal. Mach. Intell.}, vol.~39, no.~6, pp. 1137--1149, 2017.

\bibitem{R15_dai2016r}
J.~Dai, Y.~Li, K.~He, and J.~Sun, ``R-fcn: Object detection via region-based
  fully convolutional networks,'' in \emph{Advances in neural information
  processing systems}, 2016, pp. 379--387.

\bibitem{sermanet2013overfeat}
P.~Sermanet, D.~Eigen, X.~Zhang, M.~Mathieu, R.~Fergus, and Y.~LeCun,
  ``Overfeat: Integrated recognition, localization and detection using
  convolutional networks,'' in \emph{2nd International Conference on Learning
  Representations}, Y.~Bengio and Y.~LeCun, Eds., 2014.

\bibitem{redmon2016you}
J.~Redmon, S.~Divvala, R.~Girshick, and A.~Farhadi, ``You only look once:
  Unified, real-time object detection,'' in \emph{Proceedings of the IEEE
  conference on computer vision and pattern recognition}, 2016, pp. 779--788.

\bibitem{R11_liu2016ssd}
W.~Liu, D.~Anguelov, D.~Erhan, C.~Szegedy, S.~Reed, C.-Y. Fu, and A.~C. Berg,
  ``Ssd: Single shot multibox detector,'' in \emph{Proceedings of the European
  Conference on Computer Vision}.\hskip 1em plus 0.5em minus 0.4em\relax
  Springer, 2016, pp. 21--37.

\bibitem{lin2017feature}
T.-Y. Lin, P.~Doll{\'a}r, R.~Girshick, K.~He, B.~Hariharan, and S.~Belongie,
  ``Feature pyramid networks for object detection,'' in \emph{Proceedings of
  the IEEE conference on computer vision and pattern recognition}, 2017, pp.
  2117--2125.

\bibitem{R25_Lin2017Focal}
T.~Y. Lin, P.~Goyal, R.~Girshick, K.~He, and P.~Dollar, ``Focal loss for dense
  object detection,'' \emph{IEEE transactions on pattern analysis and machine
  intelligence}, vol.~PP, no.~99, pp. 2999--3007, 2017.

\bibitem{fu2017dssd}
C.~Fu, W.~Liu, A.~Ranga, A.~Tyagi, and A.~C. Berg, ``{DSSD} : Deconvolutional
  single shot detector,'' \emph{CoRR}, vol. abs/1701.06659, 2017.

\bibitem{cai2018cascade}
Z.~Cai and N.~Vasconcelos, ``Cascade r-cnn: Delving into high quality object
  detection,'' in \emph{Proceedings of the IEEE conference on computer vision
  and pattern recognition}, 2018, pp. 6154--6162.

\bibitem{chen2019hybrid}
K.~Chen, J.~Pang, J.~Wang, Y.~Xiong, X.~Li, S.~Sun, W.~Feng, Z.~Liu, J.~Shi,
  W.~Ouyang \emph{et~al.}, ``Hybrid task cascade for instance segmentation,''
  in \emph{Proceedings of the IEEE conference on computer vision and pattern
  recognition}, 2019, pp. 4974--4983.

\bibitem{zhang2018single}
S.~Zhang, L.~Wen, X.~Bian, Z.~Lei, and S.~Z. Li, ``Single-shot refinement
  neural network for object detection,'' in \emph{Proceedings of the IEEE
  conference on computer vision and pattern recognition}, 2018, pp. 4203--4212.

\bibitem{R14_Tian2019FCOS}
Z.~Tian, C.~Shen, H.~Chen, and T.~He, ``{FCOS:} fully convolutional one-stage
  object detection,'' in \emph{2019 {IEEE/CVF} International Conference on
  Computer Vision, {ICCV} 2019, Seoul, Korea (South), October 27 - November 2,
  2019}.\hskip 1em plus 0.5em minus 0.4em\relax {IEEE}, 2019, pp. 9626--9635.

\bibitem{kong2019foveabox}
\BIBentryALTinterwordspacing
T.~Kong, F.~Sun, H.~Liu, Y.~Jiang, and J.~Shi, ``Foveabox: Beyond anchor-based
  object detector,'' \emph{CoRR}, vol. abs/1904.03797, 2019. [Online].
  Available: \url{http://arxiv.org/abs/1904.03797}
\BIBentrySTDinterwordspacing

\bibitem{Yang_2019_ICCV}
Z.~Yang, S.~Liu, H.~Hu, L.~Wang, and S.~Lin, ``Reppoints: Point set
  representation for object detection,'' in \emph{2019 {IEEE/CVF} International
  Conference on Computer Vision}.\hskip 1em plus 0.5em minus 0.4em\relax
  {IEEE}, 2019, pp. 9656--9665.

\bibitem{R22_Jiang2017R2CNN}
\BIBentryALTinterwordspacing
Y.~Jiang, X.~Zhu, X.~Wang, S.~Yang, W.~Li, H.~Wang, P.~Fu, and Z.~Luo,
  ``{R2CNN:} rotational region {CNN} for orientation robust scene text
  detection,'' \emph{CoRR}, vol. abs/1706.09579, 2017. [Online]. Available:
  \url{http://arxiv.org/abs/1706.09579}
\BIBentrySTDinterwordspacing

\bibitem{R30_ma2018arbitrary}
J.~Ma, W.~Shao, H.~Ye, L.~Wang, H.~Wang, Y.~Zheng, and X.~Xue,
  ``Arbitrary-oriented scene text detection via rotation proposals,''
  \emph{IEEE Transactions on Multimedia}, vol.~20, no.~11, pp. 3111--3122,
  2018.

\bibitem{fu2018ship}
K.~Fu, Y.~Li, H.~Sun, X.~Yang, G.~Xu, Y.~Li, and X.~Sun, ``A ship rotation
  detection model in remote sensing images based on feature fusion pyramid
  network and deep reinforcement learning,'' \emph{Remote Sensing}, vol.~10,
  no.~12, p. 1922, 2018.

\bibitem{yang2018object}
X.~Yang, K.~Fu, H.~Sun, X.~Sun, M.~Yan, W.~Diao, and Z.~Guo, ``Object detection
  with head direction in remote sensing images based on rotational region
  cnn,'' in \emph{IGARSS 2018-2018 IEEE International Geoscience and Remote
  Sensing Symposium}.\hskip 1em plus 0.5em minus 0.4em\relax IEEE, 2018, pp.
  2507--2510.

\bibitem{yang2019building}
J.~Yang, L.~Ji, X.~Geng, X.~Yang, and Y.~Zhao, ``Building detection in high
  spatial resolution remote sensing imagery with the u-rotation detection
  network,'' \emph{International Journal of Remote Sensing}, vol.~40, no.~15,
  pp. 6036--6058, 2019.

\bibitem{R31_liao2018textboxes++}
M.~Liao, B.~Shi, and X.~Bai, ``Textboxes++: A single-shot oriented scene text
  detector,'' \emph{IEEE transactions on image processing}, vol.~27, no.~8, pp.
  3676--3690, 2018.

\bibitem{R27_azimi2018towards}
S.~M. Azimi, E.~Vig, R.~Bahmanyar, M.~K{\"o}rner, and P.~Reinartz, ``Towards
  multi-class object detection in unconstrained remote sensing imagery,'' in
  \emph{Asian Conference on Computer Vision}.\hskip 1em plus 0.5em minus
  0.4em\relax Springer, 2018, pp. 150--165.

\bibitem{R29_ding2018learning}
J.~Ding, N.~Xue, Y.~Long, G.~Xia, and Q.~Lu, ``Learning roi transformer for
  oriented object detection in aerial images,'' in \emph{{IEEE} Conference on
  Computer Vision and Pattern Recognition}.\hskip 1em plus 0.5em minus
  0.4em\relax Computer Vision Foundation / {IEEE}, 2019, pp. 2849--2858.

\bibitem{R28_Yang2018SCRDet}
X.~Yang, J.~Yang, J.~Yan, Y.~Zhang, T.~Zhang, Z.~Guo, X.~Sun, and K.~Fu,
  ``Scrdet: Towards more robust detection for small, cluttered and rotated
  objects,'' in \emph{2019 {IEEE/CVF} International Conference on Computer
  Vision}.\hskip 1em plus 0.5em minus 0.4em\relax {IEEE}, 2019, pp. 8231--8240.

\bibitem{R20_Yang2019R3Det}
X.~Yang, Q.~Liu, J.~Yan, and A.~Li, ``R3det: Refined single-stage detector with
  feature refinement for rotating object,'' \emph{arXiv preprint
  arXiv:1908.05612}, 2019.

\bibitem{2020Gliding}
Y.~Xu, M.~Fu, Q.~Wang, Y.~Wang, and X.~Bai, ``Gliding vertex on the horizontal
  bounding box for multi-oriented object detection,'' \emph{IEEE Transactions
  on Pattern Analysis and Machine Intelligence}, vol.~PP, no.~99, pp. 1--1,
  2020.

\bibitem{wang2020learning}
J.~Wang, W.~Yang, H.-C. Li, H.~Zhang, and G.-S. Xia, ``Learning center
  probability map for detecting objects in aerial images,'' \emph{IEEE
  Transactions on Geoscience and Remote Sensing}, 2020.

\bibitem{yang2020arbitrary}
X.~Yang, J.~Yan, and T.~He, ``On the arbitrary-oriented object detection:
  Classification based approaches revisited,'' \emph{arXiv preprint
  arXiv:2003.05597}, 2020.

\bibitem{fu2020rotation}
K.~Fu, Z.~Chang, Y.~Zhang, G.~Xu, K.~Zhang, and X.~Sun, ``Rotation-aware and
  multi-scale convolutional neural network for object detection in remote
  sensing images,'' \emph{ISPRS Journal of Photogrammetry and Remote Sensing},
  vol. 161, pp. 294--308, 2020.

\bibitem{he2016deep}
K.~He, X.~Zhang, S.~Ren, and J.~Sun, ``Deep residual learning for image
  recognition,'' in \emph{Proceedings of the IEEE conference on computer vision
  and pattern recognition}, 2016, pp. 770--778.

\bibitem{deng2009imagenet}
J.~Deng, W.~Dong, R.~Socher, L.-J. Li, K.~Li, and L.~Fei-Fei, ``Imagenet: A
  large-scale hierarchical image database,'' in \emph{Proceedings of the IEEE
  conference on computer vision and pattern recognition}.\hskip 1em plus 0.5em
  minus 0.4em\relax Ieee, 2009, pp. 248--255.

\bibitem{bao2019single}
S.~Bao, X.~Zhong, R.~Zhu, X.~Zhang, Z.~Li, and M.~Li, ``Single shot anchor
  refinement network for oriented object detection in optical remote sensing
  imagery,'' \emph{IEEE Access}, vol.~7, pp. 87\,150--87\,161, 2019.

\bibitem{ming2021optimization}
Q.~Ming, Z.~Zhou, L.~Miao, X.~Yang, and Y.~Dong, ``Optimization for oriented
  object detection via representation invariance loss,'' \emph{arXiv preprint
  arXiv:2103.11636}, 2021.

\bibitem{DBLP:journals/corr/abs-2011-09670}
\BIBentryALTinterwordspacing
X.~Yang, L.~Hou, Y.~Zhou, W.~Wang, and J.~Yan, ``Dense label encoding for
  boundary discontinuity free rotation detection,'' \emph{CoRR}, vol.
  abs/2011.09670, 2020. [Online]. Available:
  \url{https://arxiv.org/abs/2011.09670}
\BIBentrySTDinterwordspacing

\bibitem{yang2021rethinking}
X.~Yang, J.~Yan, Q.~Ming, W.~Wang, X.~Zhang, and Q.~Tian, ``Rethinking rotated
  object detection with gaussian wasserstein distance loss,'' \emph{arXiv
  preprint arXiv:2101.11952}, 2021.

\bibitem{yang2021learning}
X.~Yang, X.~Yang, J.~Yang, Q.~Ming, W.~Wang, Q.~Tian, and J.~Yan, ``Learning
  high-precision bounding box for rotated object detection via kullback-leibler
  divergence,'' \emph{arXiv preprint arXiv:2106.01883}, 2021.

\bibitem{lin2019ienet}
Y.~Lin, P.~Feng, and J.~Guan, ``Ienet: Interacting embranchment one stage
  anchor free detector for orientation aerial object detection,'' \emph{arXiv
  preprint arXiv:1912.00969}, 2019.

\bibitem{R21_yang2018automatic}
X.~Yang, H.~Sun, K.~Fu, J.~Yang, X.~Sun, M.~Yan, and Z.~Guo, ``Automatic ship
  detection in remote sensing images from google earth of complex scenes based
  on multiscale rotation dense feature pyramid networks,'' \emph{Remote
  Sensing}, vol.~10, no.~1, p. 132, 2018.

\bibitem{zhang2019cad}
G.~Zhang, S.~Lu, and W.~Zhang, ``Cad-net: A context-aware detection network for
  objects in remote sensing imagery,'' \emph{IEEE Transactions on Geoscience
  and Remote Sensing}, vol.~57, no.~12, pp. 10\,015--10\,024, 2019.

\bibitem{zhu2020adaptive}
Y.~Zhu, J.~Du, and X.~Wu, ``Adaptive period embedding for representing oriented
  objects in aerial images,'' \emph{IEEE Transactions on Geoscience and Remote
  Sensing}, vol.~58, no.~10, pp. 7247--7257, 2020.

\bibitem{zhou2020objects}
L.~Zhou, H.~Wei, H.~Li, Y.~Zhang, X.~Sun, and W.~Zhao, ``Objects detection for
  remote sensing images based on polar coordinates,'' \emph{arXiv preprint
  arXiv:2001.02988}, 2020.

\bibitem{wei2019oriented}
H.~Wei, Y.~Zhang, Z.~Chang, H.~Li, H.~Wang, and X.~Sun, ``Oriented objects as
  pairs of middle lines,'' \emph{ISPRS Journal of Photogrammetry and Remote
  Sensing}, vol. 169, pp. 268--279, 2020.

\bibitem{2020Dynamic}
X.~Pan, Y.~Ren, K.~Sheng, W.~Dong, H.~Yuan, X.~Guo, C.~Ma, and C.~Xu, ``Dynamic
  refinement network for oriented and densely packed object detection,'' in
  \emph{2020 {IEEE/CVF} Conference on Computer Vision and Pattern
  Recognition}.\hskip 1em plus 0.5em minus 0.4em\relax {IEEE}, 2020, pp.
  11\,204--11\,213.

\bibitem{he2017deep}
W.~He, X.-Y. Zhang, F.~Yin, and C.-L. Liu, ``Deep direct regression for
  multi-oriented scene text detection,'' in \emph{Proceedings of the IEEE
  conference on computer vision and pattern recognition}, 2017, pp. 745--753.

\bibitem{liu2018fots}
X.~Liu, D.~Liang, S.~Yan, D.~Chen, Y.~Qiao, and J.~Yan, ``Fots: Fast oriented
  text spotting with a unified network,'' in \emph{Proceedings of the IEEE
  conference on computer vision and pattern recognition}, 2018, pp. 5676--5685.

\bibitem{liu2017learning}
\BIBentryALTinterwordspacing
L.~Liu, Z.~Pan, and B.~Lei, ``Learning a rotation invariant detector with
  rotatable bounding box,'' \emph{CoRR}, vol. abs/1711.09405, 2017. [Online].
  Available: \url{http://arxiv.org/abs/1711.09405}
\BIBentrySTDinterwordspacing

\bibitem{li2019feature}
C.~Li, C.~Xu, Z.~Cui, D.~Wang, T.~Zhang, and J.~Yang, ``Feature-attentioned
  object detection in remote sensing imagery,'' in \emph{IEEE International
  Conference on Image Processing}.\hskip 1em plus 0.5em minus 0.4em\relax IEEE,
  2019, pp. 3886--3890.

\end{thebibliography}

\end{document}